\documentclass{article}

\usepackage{dits_arxiv,times}

\usepackage[T1]{fontenc}

\usepackage{microtype}
\usepackage{graphicx}
\usepackage{float}
\usepackage{placeins}
\usepackage{subcaption}
\usepackage{wrapfig}
\usepackage{booktabs}
\usepackage{multirow,longtable}

\usepackage{hyperref}
\usepackage{url}
\hypersetup{hidelinks}

\usepackage{amsmath}
\usepackage{amssymb}

\usepackage{xcolor}

\newcommand{\boldres}[1]{{\textbf{\textcolor{red}{#1}}}}
\newcommand{\secondres}[1]{{\underline{\textcolor{blue}{#1}}}}

\title{DiTS: Multimodal Diffusion Transformers Are Time Series Forecasters}

\author{Haoran Zhang\textsuperscript{*}, Haixuan Liu\textsuperscript{*}, Yong Liu\textsuperscript{*}, Yunzhong Qiu,\\
\textbf{Yuxuan Wang, Jianmin Wang, Mingsheng Long\textsuperscript{$\boxtimes$}}\\
\normalfont School of Software, BNRist, Tsinghua University\\
\normalfont\texttt{\{zhang-hr24,liuhaixu21,liuyong21\}@mails.tsinghua.edu.cn}\\
\normalfont\textsuperscript{$\boxtimes$}\texttt{mingsheng@tsinghua.edu.cn}}
\hypersetup{
  pdftitle={DiTS: Multimodal Diffusion Transformers Are Time Series Forecasters},
  pdfauthor={Haoran Zhang, Haixuan Liu, Yong Liu, Yunzhong Qiu, Yuxuan Wang, Jianmin Wang, Mingsheng Long},
  pdfsubject={Covariate-aware time series forecasting},
  pdfkeywords={Time Series Forecasting, Diffusion Transformer, Covariate-aware Forecasting}
}

\begin{document}
\setlength{\parskip}{4pt}

\maketitle
\begingroup
\renewcommand{\thefootnote}{}
\raggedright
\footnotetext{\textsuperscript{*}Equal Contribution}
\endgroup

\begin{abstract}

While generative modeling facilitates probabilistic time series forecasting, incorporating heterogeneous exogenous information remains challenging. Diffusion Transformers (DiT) provide a scalable framework for conditional generation, yet their adaptation to forecasting calls for conditioning mechanisms tailored to time series. Endogenous targets and exogenous covariates differ in sources, semantics, and statistical characteristics, while sharing temporal coordinates that support fine-grained conditional guidance. Covariates can describe future variability and temporal dependence beyond the conditional mean targeted by direct regression. Motivated by these considerations, we propose Diffusion Transformers for Time Series (DiTS), a Multimodal Diffusion Transformer for covariate-aware forecasting. DiTS models endogenous targets and exogenous covariates as distinct modalities, jointly conditioning future generation on target history and available covariates. Flow matching makes covariate-dependent distributional information relevant to velocity prediction conditioned on noisy future states. We introduce \emph{Time-aligned Modulation}, extending AdaLN with the temporal-alignment prior to generate patch-wise modulation parameters from aligned covariates and diffusion time. Across diverse covariate-aware forecasting tasks, DiTS achieves strong performance in both deterministic and probabilistic forecasting, demonstrating the effectiveness of conditional generation for both point and distributional forecasting.
\looseness=-1
\end{abstract}

\section{Introduction}

\begin{wrapfigure}{R}{0.55\textwidth}
\vspace{-15pt}
\centering
\includegraphics[width=\linewidth]{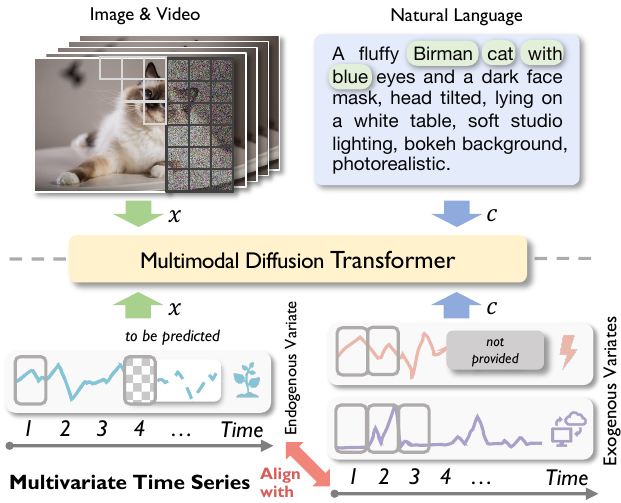}
\caption{Multimodal generation and covariate-aware forecasting. DiTS treats endogenous targets and exogenous covariates as distinct modalities, exploiting their temporal alignment for fine-grained conditional control.}
\label{fig:intro}
\vspace{-12pt}
\end{wrapfigure}

Time series forecasting is a cornerstone of decision-making, from system monitoring to energy management. The evolution of an endogenous variate is shaped by both internal temporal dependencies and diverse exogenous factors, making external context a valuable source of guidance for anticipating its future outcomes. Nevertheless, targets and covariates often originate from different sources, with distinct semantics, units, and statistical characteristics. This modality-level gap calls for preserving their individual characteristics while modeling their dependencies. Meanwhile, regularly sampled multivariate series typically share a common clock, providing a natural temporal correspondence for integrating heterogeneous information.

The development of deep forecasting models has been largely characterized by deterministic frameworks, with Transformers emerging as a prominent paradigm. Representative approaches refine temporal patch modeling~\citep{nie2022time}, represent variates as tokens~\citep{liu2023itransformer}, or distinguish endogenous modeling from exogenous aggregation~\citep{wang2024timexer}. However, providing access to covariates does not by itself determine which aspects of their information the model learns to use. Changing external conditions can shift future distributions beyond what target history suggests. Covariates help account for these shifts, but direct squared-error regression targets only their effect on the conditional mean, leaving changes in variability and temporal dependence outside its prediction target.

Recently, Diffusion Transformers~\citep{peebles2023scalable} have provided a scalable backbone for conditional generation, with multimodal variants enabling fine-grained interaction between distinct input representations~\citep{esser2024scaling}. This perspective extends to target generation guided by exogenous covariates (Figure~\ref{fig:intro}). We adopt conditional flow matching~\citep{lipman2022flow,liu2022flow} to learn a velocity field conditioned on target history and observed covariates, using probability paths from noise to future targets. At intermediate states, the optimal velocity depends on the expected clean future given the current state and these conditions. Covariate-dependent distributional differences can therefore affect velocity prediction even when the original conditional mean is unchanged, providing supervision beyond direct mean regression.

The common temporal coordinates of multivariate series offer a basis for adapting Multimodal Diffusion Transformers to forecasting. Corresponding target and covariate patches cover the same interval, providing a strong prior for where external information should enter. Rather than broadcasting a series-level representation across the window, AdaLN~\citep{peebles2023scalable} can exploit this correspondence to generate patch-specific parameters from aligned covariates and diffusion time. This motivates \emph{Time-aligned Modulation}: temporal position determines where conditioning applies, while diffusion time specifies the generation stage. Temporal attention propagates conditioned information across positions, accommodating lagged dependencies without adding covariate tokens to attention.

To this end, we propose Diffusion Transformers for Time Series (\textbf{DiTS}), a Multimodal Diffusion Transformer integrating Time-aligned Modulation with conditional flow matching. DiTS treats endogenous targets and exogenous covariates as distinct modalities with different sources, semantics, and statistical characteristics but shared temporal coordinates. Generation is jointly conditioned on target history and available covariates: target patches retain historical context, while a \emph{Covariate Encoder} preserves patch-wise external information. Time-aligned Modulation connects these representations, and DiTS blocks refine target patches through modulated temporal attention and feed-forward transformations. The learned velocity field supports probabilistic generation and sample-averaged point prediction. Experiments demonstrate strong forecasting performance, particularly with known future covariates, while ablations examine the contributions of conditional velocity learning and time-aligned injection. Our main contributions are summarized as follows:
\begin{itemize}
    \item We propose DiTS, a Multimodal Diffusion Transformer for covariate-aware forecasting, using conditional flow matching to incorporate exogenous information into forecast generation and improve predictive accuracy over direct mean regression on the evaluated datasets.
    \item We introduce Time-aligned Modulation, extending AdaLN with patch-wise parameters generated from temporally aligned covariates. With a Covariate Encoder and temporal attention, it enables fine-grained conditional control without expanding the attention sequence.
    \item DiTS achieves strong performance across diverse covariate-aware forecasting tasks, attaining leading aggregate results in both deterministic and probabilistic forecasting. Ablation studies further validate its covariate utilization and generation efficiency.
\end{itemize}

\section{Related Works}

\subsection{Forecasting with Exogenous Variables}

Forecasting with exogenous variables requires relating target history to external observations that may be available only in the past or known over the prediction horizon. General multivariate Transformers organize temporal and cross-variate dependencies through channel-independent patches in PatchTST~\citep{nie2022time}, variate tokens in iTransformer~\citep{liu2023itransformer}, separate temporal and cross-dimension interactions in Crossformer~\citep{zhang2022crossformer}, or autoregressive multivariate patches in Timer-XL~\citep{liu2024timerxl}. Methods designed for exogenous forecasting make the roles of targets and covariates more explicit. TFT~\citep{lim2021temporal} uses variable selection and attention to integrate heterogeneous inputs for multi-horizon prediction, while TimeXer~\citep{wang2024timexer} models endogenous series and aggregates exogenous information asymmetrically. DAG~\citep{qiu2025dag} discovers and injects temporal and channel correlations involving historical and future exogenous variables. GCGNet~\citep{li2026gcgnet} combines a variational generator, graph-structure alignment, and refinement to model correlations across time and variables in exogenous-aware forecasting.
These advances highlight the importance of organizing information at an appropriate granularity. For covariate-aware forecasting, however, external information can describe more than changes in the future mean. Preserving its temporal correspondence and learning its distributional implications provide complementary opportunities beyond refining tokenization or attention alone.

\subsection{Diffusion Models for Time Series}

Early diffusion approaches include TimeGrad~\citep{rasul2021autoregressive} for autoregressive forecasting and CSDI~\citep{tashiro2021csdi} for probabilistic imputation. Subsequent explorations incorporate structured state-space models in SSSD~\citep{alcaraz2022diffusion}, latent diffusion in LDT~\citep{feng2024latent}, and general-purpose diffusion Transformers in TimeDiT~\citep{cao2024timedit}, exploring representations and backbones for time series generation. Recently, Sundial~\citep{liu2025sundial} combines autoregressive context with flow-matching prediction. Beyond diffusion, TempFlow~\citep{rasul2021multivariate} uses conditioned normalizing flows for multivariate probabilistic forecasting, offering an alternative framework for capturing predictive distributions and cross-variate dependencies.

Beyond quantifying uncertainty, conditional generation makes distributional information relevant to forecasting. In flow matching, the same noisy state may require different updates under different exogenous conditions, even when their original conditional means agree. This motivates learning how covariates influence trajectories, complementing the development of expressive generative backbones.

\subsection{Diffusion Transformers}

Adopting Transformers has advanced scalability and unification in generative modeling. DiT~\citep{peebles2023scalable} introduces AdaLN-Zero to incorporate timestep and class information through adaptive normalization. PixArt-$\alpha$~\citep{chen2023pixart} adopts cross-attention for text conditioning, while Multimodal Diffusion Transformers~\citep{esser2024scaling} combine rectified flow with modality-specific representations and joint attention. These developments demonstrate the value of accommodating heterogeneous inputs while maintaining fine-grained conditional control over generation.

Their adaptation to time series can exploit the temporal coordinates shared by targets and covariates. DiTS develops Time-aligned Modulation to preserve this correspondence within AdaLN, supplying each target patch with its covariate-dependent parameters. Temporal attention carries the conditioned information across patches, combining local conditional control with flexible temporal modeling.

\section{Approach}

\begin{wrapfigure}{r}{0.55\textwidth}
\vspace{-10pt}
\centering
\includegraphics[width=\linewidth]{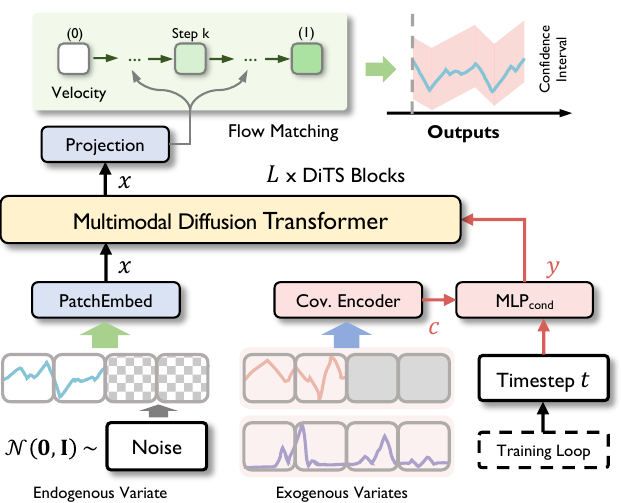}
\caption{DiTS overview. Target patches pass through $L$ DiTS blocks conditioned by aligned covariates and diffusion time via Time-aligned Modulation. The projected velocity guides forecast generation.}\label{fig:overview}
\vspace{-20pt}
\end{wrapfigure}

We introduce DiTS, a Multimodal Diffusion Transformer for covariate-aware forecasting. As illustrated in Figure~\ref{fig:overview}, target patches and a Covariate Encoder represent both modalities separately. Time-aligned Modulation injects aligned external information into temporal attention and feed-forward transformations, while conditional flow matching supports training and probabilistic forecasting.

\subsection{Preliminaries}
Let $\mathbf{x}_{0,\mathrm{his}}\in\mathbb{R}^{W}$ denote the historical observations of a target series and $\mathbf{y}_{0,\mathrm{pred}}\in\mathbb{R}^{H}$ its future values. The subscript $0$ identifies the target, and $1{:}C$ indexes the exogenous covariates. We consider forecasting with both known future and past-only covariates, using only observations available at forecast time. For clarity, the following equations describe the known-future case, with $\mathbf{x}_{1:C}\in\mathbb{R}^{T\times C}$ and $T=W+H$. DiTS learns the conditional distribution of future targets given their history and the available exogenous observations.

We adopt the flow matching paradigm, specifically rectified flow~\citep{lipman2022flow,liu2022flow}, to learn a continuous transformation from Gaussian noise to the data distribution. Given independent noise $\boldsymbol{\epsilon}\sim\mathcal{N}(\mathbf{0},\mathbf{I}_H)$, we define a linear interpolation between noise and future targets:
\begin{equation}
\mathbf{x}_{0,\mathrm{pred}}(t)
=(1-t)\boldsymbol{\epsilon}+t\mathbf{y}_{0,\mathrm{pred}},
\qquad t\in[0,1].
\label{eq:noising}
\end{equation}
This formulation defines straight paths from noise at $t=0$ to data at $t=1$. We learn a network velocity field $\mathbf{v}_{\theta}$ conditioned on target history and exogenous covariates to approximate the corresponding time-dependent vector field.

The training objective minimizes the mean squared error between the predicted velocity and the target drift. Since the path is linear, the target velocity is $\mathbf{y}_{0,\mathrm{pred}}-\boldsymbol{\epsilon}$:
\begin{equation}
\mathcal{L}_{\mathrm{velocity}}(\theta)
=\mathbb{E}\left[
\left\|
\mathbf{v}_{\theta}(\mathbf{x}_{0,\mathrm{pred}}(t),t;\mathbf{x}_{0,\mathrm{his}},\mathbf{x}_{1:C})
-(\mathbf{y}_{0,\mathrm{pred}}-\boldsymbol{\epsilon})
\right\|_2^2
\right].
\label{eq:velocity_loss}
\end{equation}
The expectation is over training windows, noise, and independently sampled diffusion times. We use logit-normal time sampling, $t=\operatorname{sigmoid}(u)$ with $u\sim\mathcal{N}(0,1)$. The path and loss use the normalized coordinates defined below; history and covariates remain observed conditions. Once trained, the model generates samples by solving the ordinary differential equation defined by the learned velocity field, integrating from Gaussian noise at $t=0$ toward $t=1$ without observing future target values.

\subsection{Time Series Tokenization}

\paragraph{Re-Normalization} To account for scale variation across forecasting windows, we apply sample-wise normalization~\citep{liu2022non}. Target history and training targets use statistics computed exclusively from $\mathbf{x}_{0,\mathrm{his}}$. Each covariate is normalized separately over its available observations: $W+H$ values for known future covariates and $W$ for past-only covariates. We use population variances and add a stabilizer $\varepsilon=10^{-5}$ before taking the square root. These statistics remain fixed during sampling. The following symbols denote normalized sequences; generated forecasts are subsequently restored to the target scale.

\paragraph{Patch Embedding} We concatenate target history and the current future state as $\mathbf{x}_0(t)=[\mathbf{x}_{0,\mathrm{his}};\mathbf{x}_{0,\mathrm{pred}}(t)]\in\mathbb{R}^{T}$. Historical observations remain fixed, while the future state follows Equation~\ref{eq:noising} during training and numerical integration during sampling. With $W$ and $H$ divisible by patch length $P$, we obtain $N=T/P$ non-overlapping patches, comprising $N_{\mathrm{his}}=W/P$ historical and $N_{\mathrm{pred}}=H/P$ future patches. A shared linear map produces their embeddings:
\begin{equation}
\mathbf{Z}_{x,n}^{0}
=\operatorname{PatchEmbed}\!\left(\mathbf{x}_0^{(n)}(t)\right)
=\mathbf{W}_{\mathrm{emb}}\mathbf{x}_0^{(n)}(t)+\mathbf{b}_{\mathrm{emb}},
\qquad n=1,\ldots,N,
\end{equation}
where $\mathbf{x}_0^{(n)}(t)\in\mathbb{R}^{P}$, $\mathbf{W}_{\mathrm{emb}}\in\mathbb{R}^{D\times P}$, and $\mathbf{b}_{\mathrm{emb}}\in\mathbb{R}^{D}$. Historical and future patches share the same embedding parameters, yielding $\mathbf{Z}_x^0\in\mathbb{R}^{N\times D}$ in temporal order.

\paragraph{Covariate Encoder} Considering the heterogeneity of endogenous and exogenous variates, we encode covariates separately while preserving temporal correspondence with target patches. As shown in Figure~\ref{fig:method}, the $C$ covariate patches at position $n$ form $\mathbf{X}_{c}^{(n)}\in\mathbb{R}^{C\times P}$ with the same boundaries as the target. Flattening in a fixed variate order and applying two shared linear layers gives
\begin{equation}
\mathbf{c}_n
=\operatorname{Linear}_{c,2}\!\left(
\operatorname{SiLU}\!\left(
\operatorname{Linear}_{c,1}\!\left(
\operatorname{Flatten}(\mathbf{X}_{c}^{(n)})
\right)\right)\right).
\label{eq:covariate_encoder}
\end{equation}
This yields $\mathbf{c}=[\mathbf{c}_1;\ldots;\mathbf{c}_N]\in\mathbb{R}^{N\times D}$, retaining a separate condition for each temporal patch. The encoder shares weights across positions, with fixed variate order preserving covariate identities. Its output is reused across DiTS blocks and integration steps; diffusion-time embedding and modulation-parameter generation are performed by Time-aligned Modulation.
When future covariates are unavailable, we first encode the historical covariate patches and use a shared linear adapter along the temporal axis to map their representations to the future patch positions. Concatenating these projected conditions with the historical conditions yields the same aligned sequence of $N$ covariate representations; all subsequent DiTS operations are identical to the known-future case.

\subsection{DiTS Block}
\label{sec:dits_block}

\begin{figure}[t]
\centering
\includegraphics[width=\textwidth]{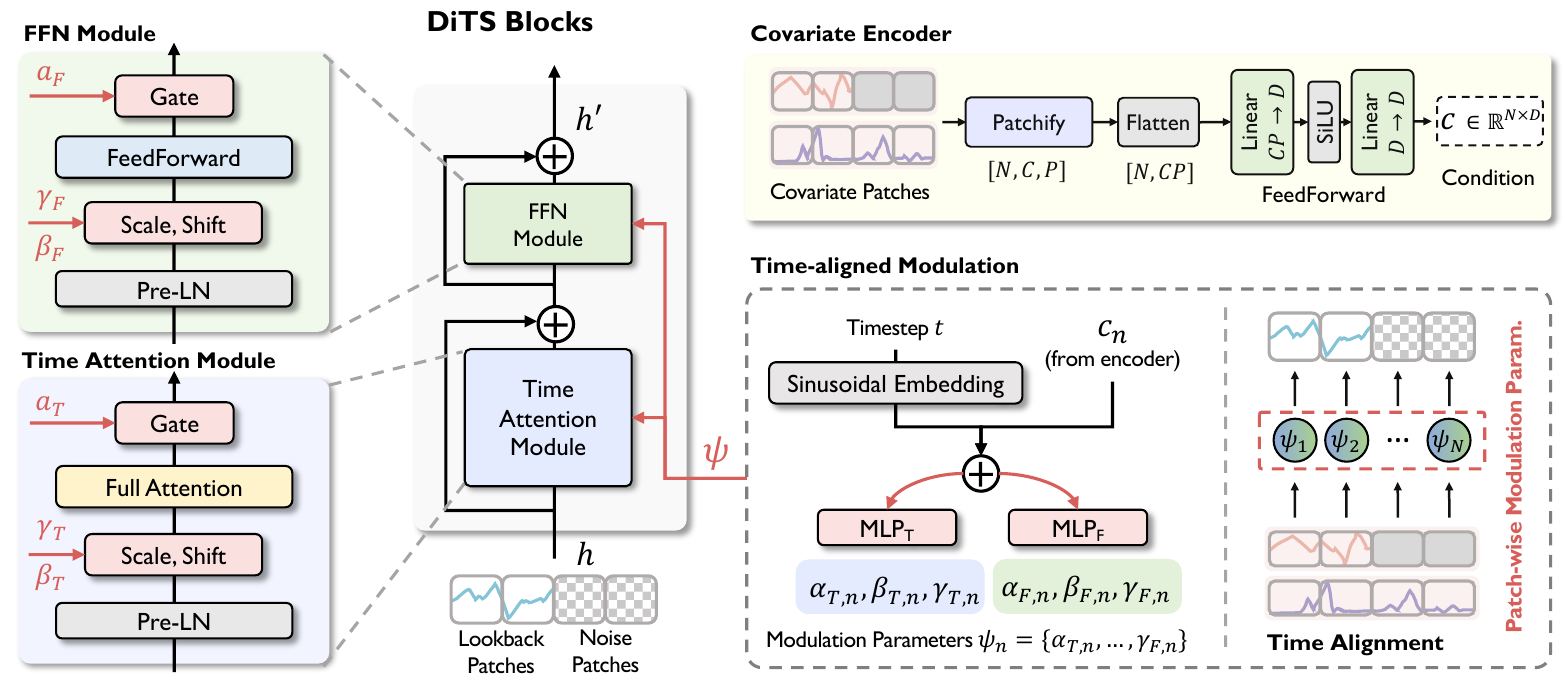}
\caption{DiTS blocks and Time-aligned Modulation. Aligned covariate conditions $\mathbf{c}_n$ combine with diffusion time to generate patch-wise parameters $\psi_n$ for scale, shift, and residual gates. This exploits temporal correspondence for local conditional control rather than broadcasting a single global condition across the window, while Time Attention propagates information across target patches.}
\label{fig:method}
\vspace{-10pt}
\end{figure}

The DiTS block combines temporal modeling and covariate injection through Time-aligned Modulation (Figure~\ref{fig:method}). It contains temporal attention and a point-wise FFN, both guided by aligned covariate conditions. Its input $h$ and output $h'$ denote the target representations $\mathbf{Z}_x^{l-1}$ and $\mathbf{Z}_x^l$.

\paragraph{Time-aligned Modulation} Target and covariate patches at position $n$ cover the same interval. Exploiting this prior, each block generates patch-wise modulation parameters from $\mathbf{c}_n$ and the diffusion-time embedding $\mathbf{e}_{\mathrm{time}}(t)\in\mathbb{R}^{D}$, obtained by applying an MLP to a sinusoidal encoding of $t$:
\begin{equation}
\left[\boldsymbol{\alpha}_{m,n}^{l}(t),\boldsymbol{\beta}_{m,n}^{l}(t),\boldsymbol{\gamma}_{m,n}^{l}(t)\right]
=\operatorname{MLP}_{m}^{l}\!\left(\mathbf{c}_n+\mathbf{e}_{\mathrm{time}}(t)\right),
\quad m\in\{T,F\}.
\label{eq:local_modulation}
\end{equation}
The $D$-dimensional vectors $\boldsymbol{\alpha}$, $\boldsymbol{\beta}$, and $\boldsymbol{\gamma}$ are the residual gate, shift, and scale offset for temporal attention ($T$) or the FFN ($F$). We collect them as
\begin{equation}
\psi_n^l(t)=
\left\{\boldsymbol{\alpha}_{T,n}^{l}(t),\boldsymbol{\beta}_{T,n}^{l}(t),\boldsymbol{\gamma}_{T,n}^{l}(t),
\boldsymbol{\alpha}_{F,n}^{l}(t),\boldsymbol{\beta}_{F,n}^{l}(t),\boldsymbol{\gamma}_{F,n}^{l}(t)\right\}.
\label{eq:psi}
\end{equation}
We write $\psi_n$ when $l,t$ are implicit and collect the $N$ sets as $\Psi^l(t)=(\psi_1^l(t),\ldots,\psi_N^l(t))$. \emph{Patch-wise Modulation Parameters} denotes separately generated values at each position, using weights shared across positions within each block. The two heads are implemented together by SiLU and a zero-initialized $D\rightarrow6D$ projection. Consequently, modulation responds to both the local covariate context and the current generation stage, preserving the distinct roles of sequence time and diffusion time.

For $\mathbf{H}\in\mathbb{R}^{N\times D}$, modulation acts on each aligned patch:
\begin{equation}
[\operatorname{Mod}_{m}^{l}(\mathbf{H})]_n
=(\mathbf{1}+\boldsymbol{\gamma}_{m,n}^{l}(t))\odot\operatorname{LN}_{m}^{l}(\mathbf{H}_n)
+\boldsymbol{\beta}_{m,n}^{l}(t),
\label{eq:time_aligned_modulation}
\end{equation}
with non-affine LayerNorm following AdaLN-Zero~\citep{peebles2023scalable}. The scale and shift adapt the normalized features according to each patch's temporally aligned covariate context, while the residual gate controls the module's contribution. Zero-initialized gates make each block an identity map at initialization. Below, $t$ is implicit and $\boldsymbol{\alpha}_m^l\in\mathbb{R}^{N\times D}$ stacks the patch-wise gates.

\paragraph{Time Attention and FFN} Each block applies temporal attention followed by a point-wise FFN, with Time-aligned Modulation conditioning both transformations. Patch-wise residual gates control their contributions to the target representations:
\begin{equation}
\begin{split}
\hat{\mathbf{Z}}_x^l
&=\mathbf{Z}_x^{l-1}+\boldsymbol{\alpha}_T^l\odot
\operatorname{MSA}_{\mathrm{time}}^l\!\left(\operatorname{Mod}_T^l(\mathbf{Z}_x^{l-1})\right),\\
\mathbf{Z}_x^l
&=\hat{\mathbf{Z}}_x^l+\boldsymbol{\alpha}_F^l\odot
\operatorname{FFN}^l\!\left(\operatorname{Mod}_F^l(\hat{\mathbf{Z}}_x^l)\right).
\end{split}
\end{equation}
Temporal attention uses Q/K RMS normalization and rotary positional embeddings, with full connectivity over historical and noisy future patches. All future coordinates are updated jointly, without causal masking. Since modulation precedes attention, covariates can influence temporal aggregation beyond their aligned positions, accommodating lagged dependencies. The FFN applies $D\rightarrow D_{\mathrm{ff}}\rightarrow D$ linear maps with GELU to each token. Only target representations evolve through the blocks; covariates supply modulation conditions. This keeps attention over $N$ target tokens, with pairwise cost $\mathcal{O}(N^2D)$, while the Covariate Encoder's input projection grows linearly with the number of covariates. Fine-grained conditions therefore do not require attention over an expanded time--variate token grid.

\subsection{DiTS Backbone}

The backbone stacks $L$ DiTS blocks, each conditioned by the modulation parameter sets $\Psi^l(t)$:
\begin{equation}
\mathbf{Z}_x^l=\operatorname{DiTS\text{-}Block}_l\!\left(\mathbf{Z}_x^{l-1},\Psi^l(t)\right),
\qquad l=1,\ldots,L.
\end{equation}
The Covariate Encoder output $\mathbf{c}$ is shared across blocks, while each block generates its own modulation parameters. Successive layers refine target representations through the interaction of aligned external conditions and temporal context. A zero-initialized linear head then projects the final future tokens to patch-wise velocities:
\begin{equation}
\mathbf{v}_{\theta}=\operatorname{Unpatchify}\!\left(\operatorname{Linear}_{\mathrm{out}}\!\left(\mathbf{Z}_{x,N_{\mathrm{his}}+1:N}^{L}\right)\right)\in\mathbb{R}^{H}.
\end{equation}
Historical tokens provide context, and the head predicts velocities only for the future window. Training follows Equation~\ref{eq:velocity_loss}, allowing covariates to guide state-dependent updates throughout generation.

At inference, $K$ forward Euler steps integrate the learned velocity from noise to data. With $t_k=k/K$ and $\widehat{\mathbf{x}}_{0,\mathrm{pred}}^{(0)}\sim\mathcal{N}(\mathbf{0},\mathbf{I}_H)$, the numerical state is updated by
\begin{equation}
\widehat{\mathbf{x}}_{0,\mathrm{pred}}^{(k+1)}
=\widehat{\mathbf{x}}_{0,\mathrm{pred}}^{(k)}
+\frac{1}{K}\mathbf{v}_{\theta}\!\left(
\widehat{\mathbf{x}}_{0,\mathrm{pred}}^{(k)},t_k;
\mathbf{x}_{0,\mathrm{his}},\mathbf{x}_{1:C}\right),
\quad k=0,\ldots,K-1.
\label{eq:euler_sampling}
\end{equation}
Here, $(k)$ indexes integration steps. Each step reuses the fixed history and covariate encoding, while recomputing future patch embeddings and time-dependent modulation parameters. Historical hidden representations are recomputed within each backbone evaluation and are not part of the integration state. Starting from independent noise draws, we generate $S$ trajectories, restore their original scale, and use their average for point forecasting. The samples also characterize predictive uncertainty.

\section{Experiments}

We extensively evaluate DiTS on covariate-aware forecasting, considering both point accuracy and probabilistic forecast quality. Experiments cover five electricity price forecasting (EPF) datasets and 13 selected FEV-Bench tasks, followed by analyses of covariate utilization, condition control, and generation efficiency. We report MSE, MAE, and MASE for point forecasts, and WQL and CRPS for probabilistic evaluation; lower values are better. Dataset descriptions, implementation details, and full results are provided in Appendices~\ref{appendix:datasets}, \ref{sec:implementation}, and~\ref{sec:full_results}.

\subsection{Time Series Covariate-aware Forecasting}
\label{exp:covariate}

We evaluate DiTS on complementary deterministic and probabilistic forecasting tasks, examining both point accuracy and predictive distribution quality. Together, these evaluations assess covariate-conditioned generation across different information-availability settings, application domains, and temporal resolutions, with particular emphasis on the effective utilization of exogenous information.

\vspace{-5pt}
\paragraph{Deterministic Forecasting} We evaluate electricity price forecasting on five datasets---NP, PJM, BE, FR, and DE---using external forecasts as covariates~\citep{lago2021forecasting}. Generated trajectories are averaged for point prediction. With known future covariates, DiTS achieves the lowest MSE and MAE on each dataset when averaged over $(W,H)=(168,24)$ and $(720,360)$ (Table~\ref{tab:epf_wi_future}). Average MSE and MAE improve by more than 10\% and 9.0\%, respectively, over the strongest baseline for each metric, supporting conditional generation with temporally aligned exogenous guidance.

With past-only covariates at $(W,H)=(168,24)$, DiTS retains the lowest average MSE and leads both metrics on NP, PJM, and DE (Table~\ref{tab:epf_wo_future}). Its average MAE remains competitive, although other methods lead on BE and FR. These results demonstrate effective point forecasting even when external information is restricted to history.

\begin{table}[t]
\vspace{5pt}
\caption{Deterministic EPF forecasting with known future covariates, averaged over $H=24$ and $H=360$. Lower is better; \boldres{bold} and \secondres{underlined} mark the best and second-best results. GCGNet results are from~\citet{li2026gcgnet}, and DAG results are from~\citet{qiu2025dag}.}
\vspace{-10pt}
\label{tab:epf_wi_future}
\vskip 0.15in
\begin{center}
\begin{small}
\begin{sc}
\setlength{\tabcolsep}{2.8pt}
\resizebox{\textwidth}{!}{\begin{tabular}{l|cc|cc|cc|cc|cc|cc|cc|cc}
\toprule
Model & \multicolumn{2}{c}{DiTS} & \multicolumn{2}{c}{GCGNet} & \multicolumn{2}{c}{TimeXer} & \multicolumn{2}{c}{DAG} & \multicolumn{2}{c}{iTrans.} & \multicolumn{2}{c}{Cross.} & \multicolumn{2}{c}{DUET} & \multicolumn{2}{c}{TFT} \\
\cmidrule(lr){2-3}\cmidrule(lr){4-5}\cmidrule(lr){6-7}\cmidrule(lr){8-9}\cmidrule(lr){10-11}\cmidrule(lr){12-13}\cmidrule(lr){14-15}\cmidrule(lr){16-17}  
Metric & MSE & MAE & MSE & MAE & MSE & MAE & MSE & MAE & MSE & MAE & MSE & MAE & MSE & MAE & MSE & MAE \\
\midrule
NP  & \boldres{0.290} & \boldres{0.315} & 0.346 & 0.337 & 0.375 & 0.355 & 0.362 & 0.344 & 0.354 & 0.373 & \secondres{0.300} & \secondres{0.323} & 0.411 & 0.408 & 0.379 & 0.375 \\
PJM & \boldres{0.085} & \boldres{0.171} & \secondres{0.093} & 0.186 & 0.124 & 0.225 & \secondres{0.093} & \secondres{0.180} & 0.098 & 0.197 & 0.119 & 0.212 & 0.102 & 0.197 & 0.114 & 0.207 \\
BE  & \boldres{0.375} & \boldres{0.261} & 0.423 & 0.287 & 0.432 & 0.279 & 0.423 & 0.279 & 0.513 & 0.345 & \secondres{0.403} & \secondres{0.268} & 0.515 & 0.354 & 0.454 & 0.291 \\
FR  & \boldres{0.332} & \boldres{0.202} & \secondres{0.405} & 0.230 & 0.415 & 0.230 & 0.414 & \secondres{0.219} & 0.484 & 0.261 & 0.412 & 0.232 & 0.496 & 0.327 & 0.504 & 0.257 \\
DE  & \boldres{0.270} & \boldres{0.317} & 0.387 & 0.387 & 0.406 & 0.391 & 0.370 & \secondres{0.370} & 0.408 & 0.407 & \secondres{0.356} & 0.376 & 0.482 & 0.430 & 0.489 & 0.446 \\
\midrule
AVG & \boldres{0.270} & \boldres{0.253} & 0.331 & 0.285 & 0.350 & 0.296 & 0.332 & \secondres{0.278} & 0.371 & 0.317 & \secondres{0.318} & 0.283 & 0.401 & 0.343 & 0.388 & 0.315 \\
\bottomrule
\end{tabular}}
\vspace{-10pt}
\end{sc}
\end{small}
\end{center}
\end{table}

\begin{table}[t]
\vspace{-5pt}
\caption{Deterministic EPF forecasting without future covariates at $(W=168,H=24)$; \boldres{bold} and \secondres{underlined} mark the best and second-best results. GCGNet and DAG results are from~\citet{li2026gcgnet} and~\citet{qiu2025dag}, respectively; the remaining baselines are from~\citet{wang2024timexer}.}
\vspace{-10pt}
\label{tab:epf_wo_future}
\vskip 0.15in
\begin{center}
\begin{small}
\begin{sc}
\setlength{\tabcolsep}{2.8pt}
\resizebox{\textwidth}{!}{\begin{tabular}{l|cc|cc|cc|cc|cc|cc|cc|cc}
\toprule
Model & \multicolumn{2}{c}{DiTS} & \multicolumn{2}{c}{GCGNet} & \multicolumn{2}{c}{TimeXer} & \multicolumn{2}{c}{DAG} & \multicolumn{2}{c}{iTrans.} & \multicolumn{2}{c}{PatchTST} & \multicolumn{2}{c}{Cross.} & \multicolumn{2}{c}{DLinear} \\
\cmidrule(lr){2-3}\cmidrule(lr){4-5}\cmidrule(lr){6-7}\cmidrule(lr){8-9}\cmidrule(lr){10-11}\cmidrule(lr){12-13}\cmidrule(lr){14-15}\cmidrule(lr){16-17}  
Metric & MSE & MAE & MSE & MAE & MSE & MAE & MSE & MAE & MSE & MAE & MSE & MAE & MSE & MAE & MSE & MAE \\
\midrule
NP  & \boldres{0.218} & \boldres{0.257} & \secondres{0.234} & \secondres{0.259} & 0.236 & 0.268 & 0.237 & \secondres{0.259} & 0.265 & 0.300 & 0.267 & 0.284 & 0.240 & 0.285 & 0.309 & 0.321 \\
PJM & \boldres{0.069} & \boldres{0.161} & 0.073 & 0.168 & 0.093 & 0.192 & \secondres{0.072} & \secondres{0.162} & 0.097 & 0.197 & 0.106 & 0.209 & 0.101 & 0.199 & 0.108 & 0.215 \\
BE  & \secondres{0.370} & 0.263 & \boldres{0.363} & 0.251 & 0.379 & \boldres{0.243} & 0.376 & \secondres{0.244} & 0.394 & 0.270 & 0.400 & 0.262 & 0.420 & 0.290 & 0.463 & 0.313 \\
FR  & \secondres{0.341} & 0.198 & \boldres{0.338} & \secondres{0.194} & 0.385 & 0.208 & 0.345 & \boldres{0.190} & 0.439 & 0.233 & 0.411 & 0.220 & 0.434 & 0.208 & 0.429 & 0.260 \\
DE  & \boldres{0.403} & \boldres{0.397} & \secondres{0.415} & \secondres{0.400} & 0.440 & 0.415 & 0.422 & 0.411 & 0.479 & 0.443 & 0.461 & 0.432 & 0.574 & 0.430 & 0.520 & 0.463 \\
\midrule
AVG & \boldres{0.280} & 0.255 & \secondres{0.285} & \secondres{0.254} & 0.307 & 0.265 & 0.290 & \boldres{0.253} & 0.335 & 0.289 & 0.330 & 0.282 & 0.354 & 0.282 & 0.366 & 0.314 \\
\bottomrule
\end{tabular}}
\vspace{-10pt}
\end{sc}
\end{small}
\end{center}
\end{table}

\vspace{-10pt}

\begin{figure}[!htbp]
\centering
\includegraphics[width=\textwidth]{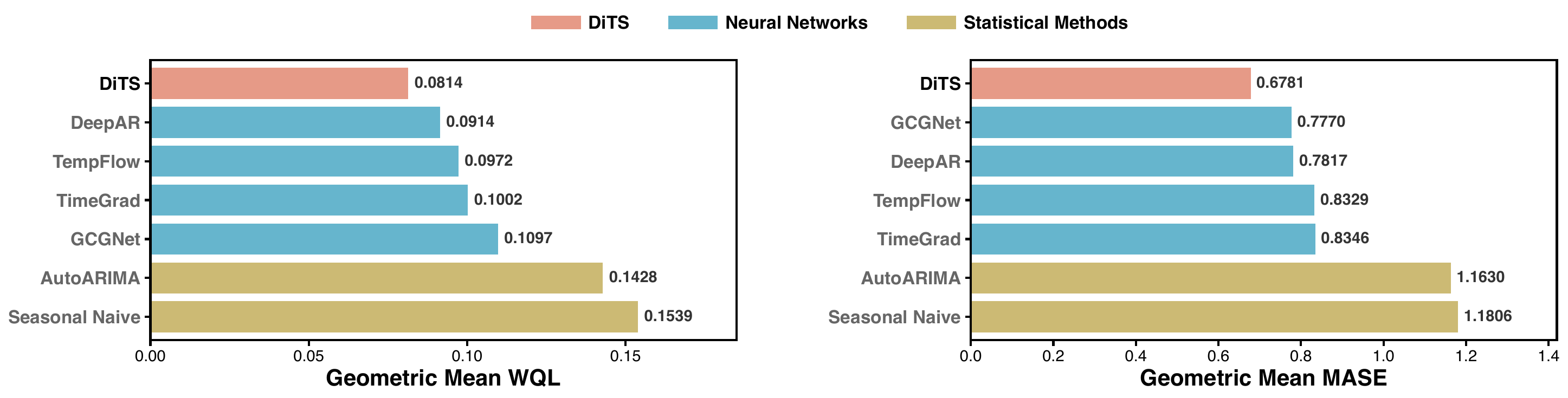}
\caption{Probabilistic forecasting on 13 covariate datasets across domains and temporal resolutions. DiTS achieves the lowest geometric mean WQL and MASE in the reported comparison, supporting covariate-conditioned generation for both distributional modeling and point forecasting.}
\label{fig:fev}
\end{figure}

\paragraph{Probabilistic Forecasting} We evaluate 13 tasks from \citet{shchur2025fev}, spanning electricity demand, prices, and solar generation. Baselines include DeepAR~\citep{Flunkert2017DeepARPF}, TimeGrad~\citep{rasul2021autoregressive}, TempFlow~\citep{rasul2021multivariate}, and GCGNet~\citep{li2026gcgnet}, alongside AutoARIMA~\citep{hyndman2008automatic} and Seasonal Naive~\citep{hyndman2018forecasting}. All neural models receive known future covariates; DiTS uses sample medians and quantiles for point and probabilistic evaluation.

DiTS achieves the lowest geometric mean WQL and MASE, improving on the strongest competing aggregates by 10.9\% and 12.7\%, respectively (Figure~\ref{fig:fev}). It performs particularly well on GFC tasks, while other models remain competitive on load and solar tasks (Table~\ref{tab:fev_full}). These results support modeling covariate-dependent distributions while retaining accurate point forecasts, consistent with richer covariate supervision through noisy-state conditioning.

\subsection{Model Analysis}
\label{sec:analysis}
\label{sec:detailed_analysis}

\paragraph{Conditional Flow Matching}
\label{exp:flowworks}

\begin{wrapfigure}{r}{0.50\linewidth}
\centering
\includegraphics[width=\linewidth]{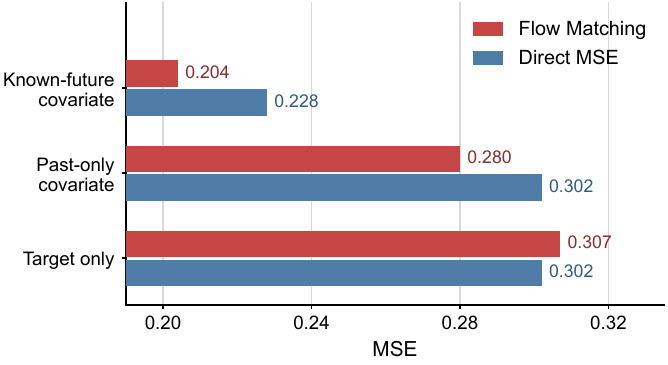}
\vspace{-20pt}
\caption{Flow matching versus direct MSE regression under three covariate settings. Bars report mean EPF MSE. Gains emerge with covariates, consistent with noisy-state conditioning making covariate-dependent distributional information relevant to velocity prediction.}
\label{fig:flow_covariate_mse}
\vspace{-10pt}
\end{wrapfigure}

We compare flow matching with direct MSE regression on EPF at $(W,H)=(168,24)$, using the same backbone, architecture hyperparameters, and optimization protocol. Flow matching improves average MSE by 10.5\% with known future covariates and 7.3\% with past-only covariates, with gains across all five datasets (Figure~\ref{fig:flow_covariate_mse}, Table~\ref{tab:flow_covariate_ablation}). Without covariates, it performs slightly worse than direct regression in this comparison. Relative to each method's target-only baseline, flow matching therefore benefits more from both types of covariates; past-only covariates provide no average improvement under direct regression at the reported precision.

These incremental gains are consistent with our analysis: given target history, covariates may convey distributional information beyond the conditional mean. Noisy-state conditioning can turn these differences into posterior-mean differences relevant to velocity prediction. Ignoring this information can increase flow-matching loss even when direct regression is unaffected, providing supervision that may improve covariate utilization.

\paragraph{Condition Control} 
\label{exp:condition}

\begin{figure}[t]
\centering
\begin{subfigure}[b]{0.48\textwidth}
    \centering
    \includegraphics[width=\linewidth]{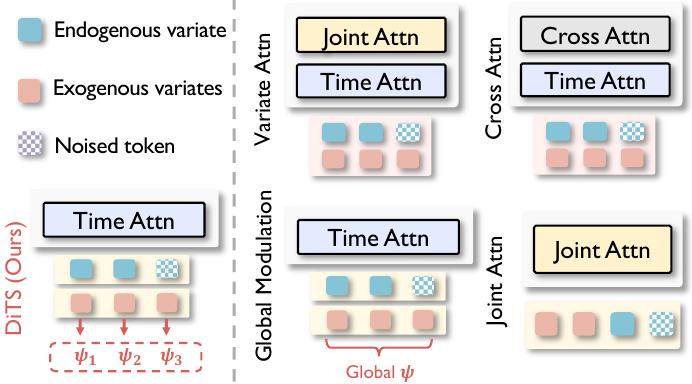}
    \caption{Covariate-conditioning architectures.}
    \label{fig:covariate_structures}
\end{subfigure}
\begin{subfigure}[b]{0.48\textwidth}
    \centering
    \includegraphics[width=\linewidth]{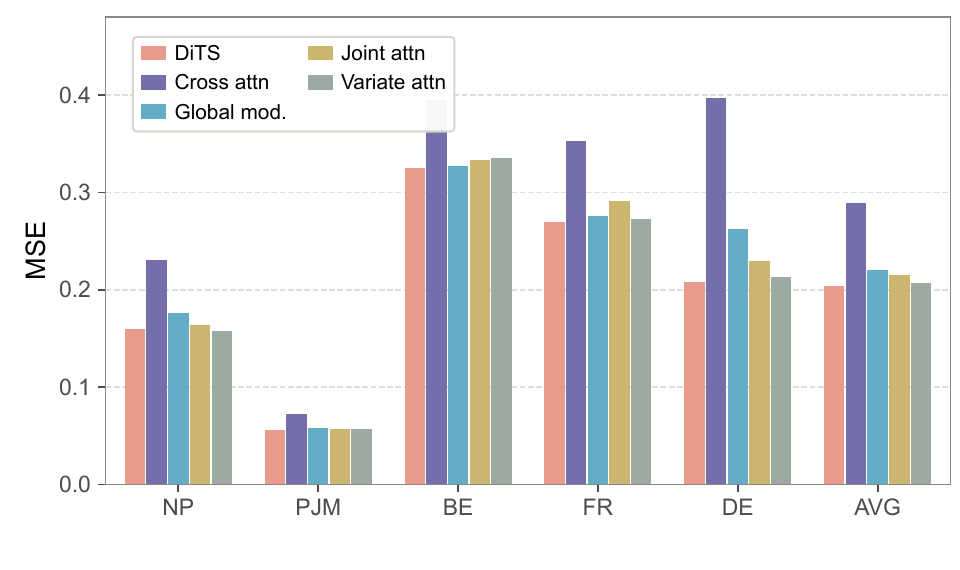}
    \caption{MSE across the EPF datasets.}
    \label{fig:covariate_results}
\end{subfigure}
\caption{Covariate-conditioning mechanisms and forecasting accuracy under $(W,H)=(168,24)$. DiTS uses Time-aligned Modulation; the alternatives use cross attention, global modulation, joint attention, or variate attention. In every variant, diffusion time enters through AdaLN; this pathway is omitted from the diagrams for clarity. The right panel reports MSE for each EPF dataset and the five-dataset average. These results favor temporally aligned modulation over global conditioning.}
\label{fig:covariate_conditioning}
\vspace{-10pt}
\end{figure}

We compare Time-aligned Modulation with cross attention, global modulation, joint attention, and variate attention (Figure~\ref{fig:covariate_conditioning}). Under $(W,H)=(168,24)$, it achieves the lowest average MSE/MAE of 0.204/0.207 (Table~\ref{tab:ablation_cond_appendix}). Relative to global modulation, average MSE decreases from 0.220 to 0.204, a 7.3\% improvement, with gains in both metrics across all five datasets. Joint, variate, and cross attention obtain MSEs of 0.215, 0.207, and 0.289, respectively.

These results support exploiting temporal correspondence in covariate injection. Time-aligned Modulation supplies external guidance at matching target positions, while temporal attention propagates its influence across the window to accommodate cross-time dependencies. Although individual datasets favor different alternatives, DiTS obtains the lowest five-dataset average MSE and MAE among these conditioning mechanisms without a separate evolving covariate stream.

\begin{figure}[!htbp]
\centering
\begin{minipage}[t]{0.49\linewidth}
\vspace{0pt}
\centering
\includegraphics[width=\linewidth]{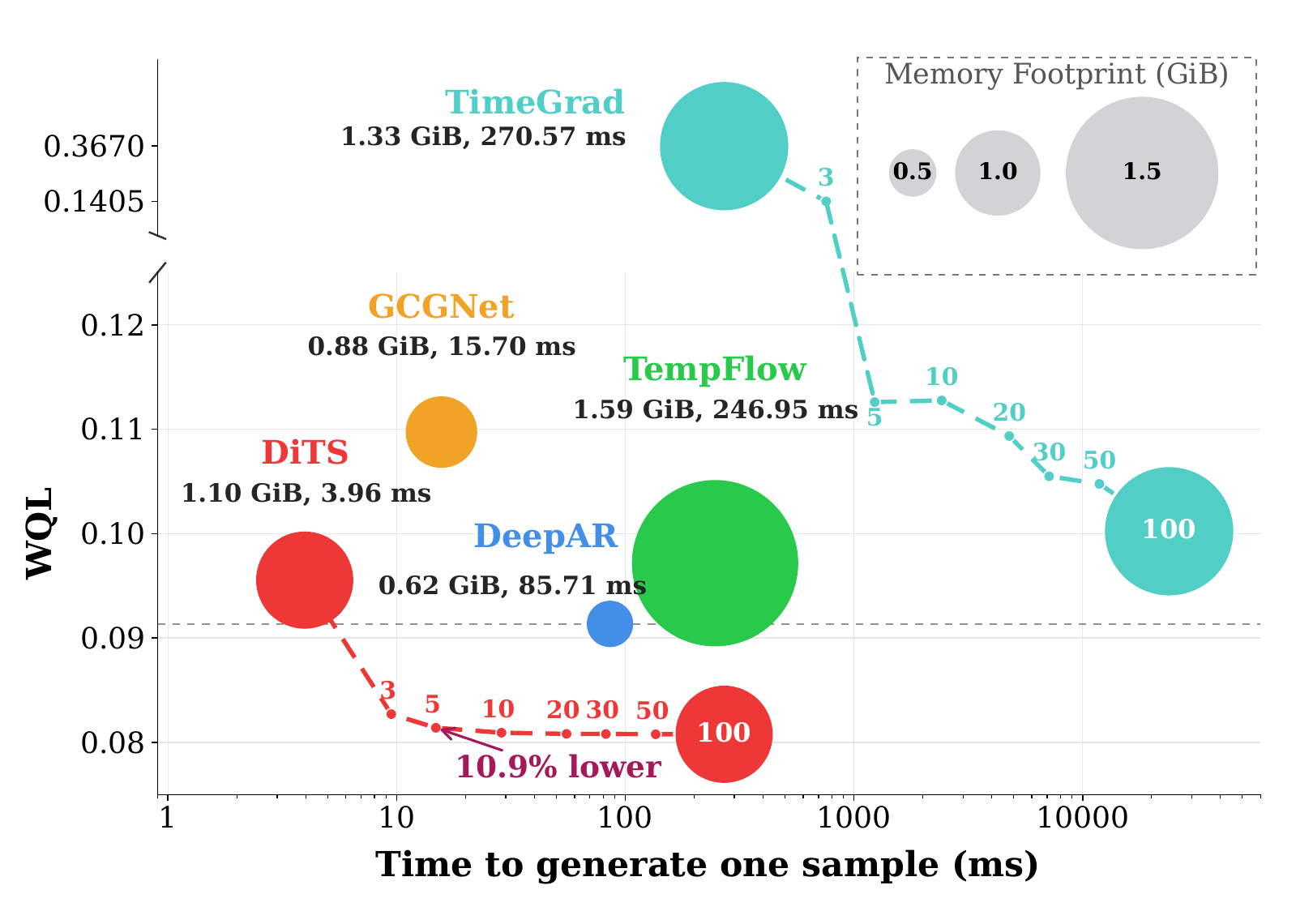}
\caption{WQL versus time on 13 datasets. Bubble area shows training memory; dashed curves vary the number of sampling steps per trajectory.}
\label{fig:fev_efficiency_tradeoff}
\end{minipage}\hfill
\begin{minipage}[t]{0.49\linewidth}
\vspace{0pt}
\centering
\includegraphics[width=\linewidth]{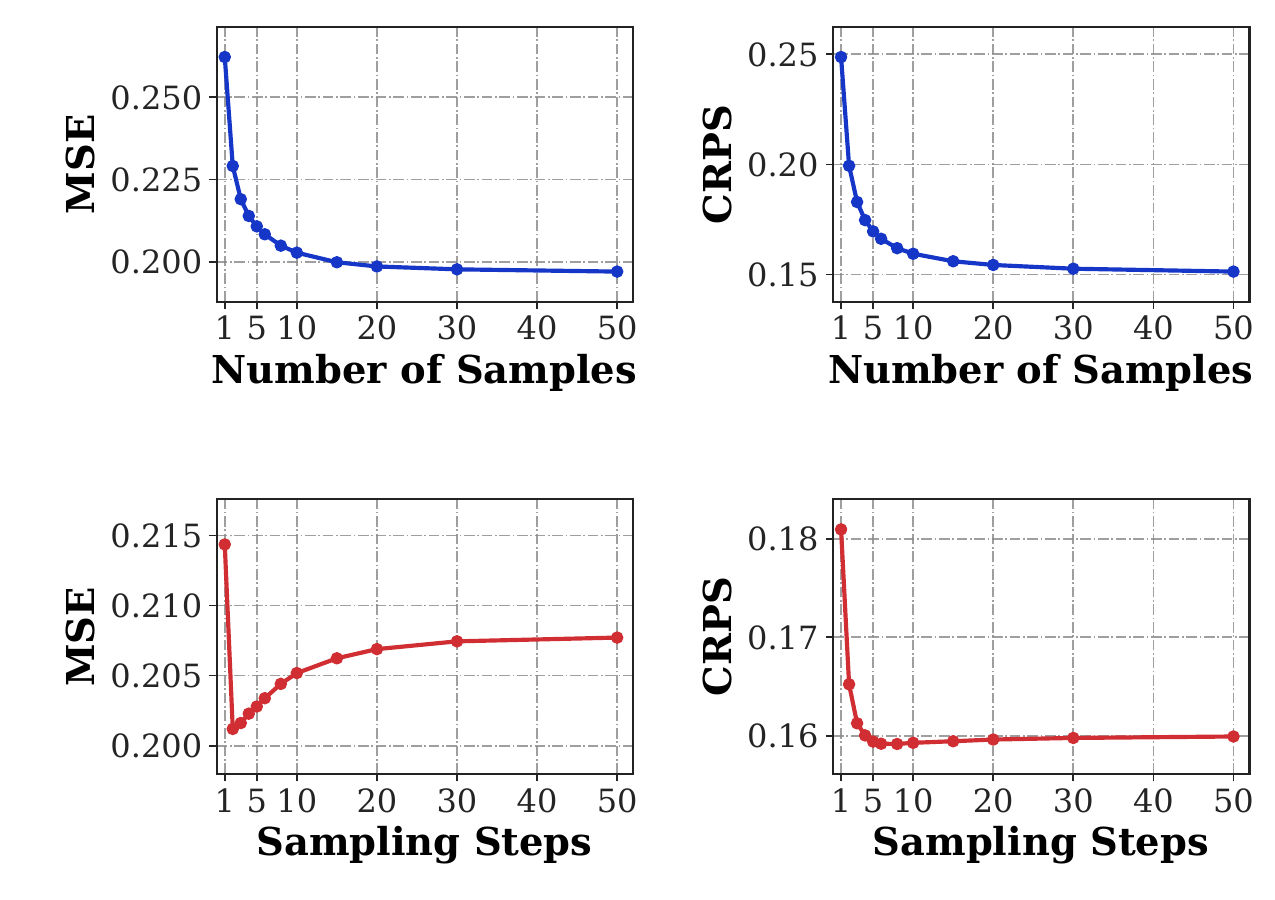}
\caption{Sample-count (top) and sampling step (bottom) sensitivity on five EPF datasets, fixing five steps and ten samples, respectively.}
\label{fig:sampling_sensitivity}
\end{minipage}
\vspace{-7pt}
\end{figure}

\paragraph{Model Efficiency} \label{exp:efficiency}
Figure~\ref{fig:fev_efficiency_tradeoff} compares WQL, generation latency, and training memory across 13 forecasting tasks, with labels indicating steps per trajectory. DiTS surpasses the baselines in WQL with three steps; further steps yield diminishing gains at higher latency. Its training memory lies below TimeGrad and TempFlow but above DeepAR and GCGNet. DiTS combines accurate forecasting with efficient few-step generation, balancing predictive performance and computational cost.

\subsection{Sampling and Noise Scheduling}
\vspace{8pt}
\label{sec:infer}

\begin{wrapfigure}{r}{0.48\linewidth}
\vspace{-8pt}
\centering
\includegraphics[width=\linewidth]{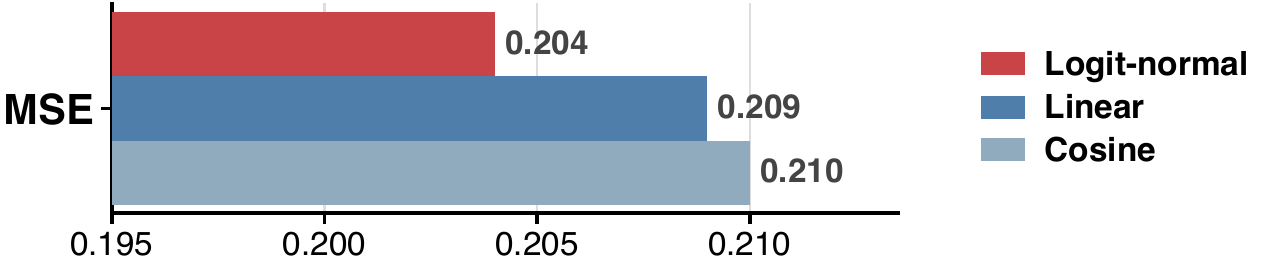}
\caption{Training-time sampling on five EPF datasets. Logit-normal gives the lowest mean MSE among the selected configurations.}
\label{fig:noise_schedule}
\vspace{-8pt}
\end{wrapfigure}

\paragraph{Inference Budget} Figure~\ref{fig:sampling_sensitivity} varies sample count $S$ with five Euler steps and integration steps $K$ with ten samples, using fixed checkpoints. More samples improve MSE and CRPS with diminishing returns. More steps initially improve CRPS before saturation, whereas MSE rises after its minimum. Higher-order solvers provide no improvement at five steps (Appendix~\ref{sec:numerical_solvers}), favoring few steps and multiple trajectories. These comparisons distinguish two uses of computation: sampling stabilizes predictive statistics, while finer integration follows the learned velocity field more closely without directly optimizing point-forecast error.

\vspace{8pt}
\paragraph{Training-Time Noise Schedule}
Training-time sampling determines how supervision is distributed across generation stages. We compare logit-normal, linear, and cosine schedules at $(W,H)=(168,24)$, obtaining mean MSE of 0.204, 0.209, and 0.210 (Figure~\ref{fig:noise_schedule}). Logit-normal leads on four datasets, while cosine leads on BE. The results favor logit-normal within this search, although the selected hyperparameters can contribute to the differences. Further comparisons of prediction parameterizations, noise paths, and guidance appear in Appendices~\ref{sec:prediction_objectives}--\ref{sec:cfg_guidance}.

\FloatBarrier

\vspace{8pt}
\section{Conclusion}
\label{sec:conclusion}

We introduced DiTS, a Multimodal Diffusion Transformer for time series generation conditioned on target history and covariates. Conditional flow matching makes covariate-dependent distributional information relevant to velocity prediction, providing supervision beyond direct mean regression. Time-aligned Modulation extends AdaLN with patch-wise parameters from aligned covariates and diffusion time, preserving temporal correspondence while temporal attention captures cross-time dependencies. Experiments demonstrate strong point and probabilistic forecasting, with ablations supporting covariate utilization under flow matching and the effectiveness of aligned modulation. DiTS achieves favorable quality with few sampling steps, balancing accuracy and generation cost. These findings support time-aligned conditional generation for covariate-aware forecasting.

\clearpage
\subsection*{AI use statement}

Generative AI tools were used solely to assist with translation, drafting, and revising portions of the manuscript, language polishing, literature search, reference formatting, and figure preparation. They were not used to develop theoretical models or conceptual frameworks, design research methodology or experiments, assist in writing proofs, implement experimental methods, generate or process datasets, support data analysis, or interpret results, nor were substantive AI-generated suggestions for mathematical assumptions, derivations, proofs, or research hypotheses adopted. All AI-assisted work was reviewed by the authors, who consulted original references, checked mathematical derivations and experimental code, and verified reported results against experimental records, and we take full responsibility for the final content of this work.

\bibliography{references}
\bibliographystyle{dits_arxiv}

\clearpage
\appendix

\section{Datasets}
\label{appendix:datasets}

This section describes the Electricity Price Forecasting (EPF) datasets and 13 probabilistic forecasting tasks selected from~\citet{shchur2025fev}, summarized in Tables~\ref{tab:epf_summary} and~\ref{tab:fev_summary}, respectively.

\subsection{Probabilistic Forecasting Task Selection}
We select tasks suitable for evaluating probabilistic forecasts with known-future covariates according to two criteria:

\begin{itemize}
    \item \textbf{Known-future covariates:} Each task provides dynamic exogenous variates available over the forecast horizon.
    \item \textbf{Sufficient historical coverage:} Each task has enough history for end-to-end training and the rolling evaluation protocol under the computational budget of this study.
\end{itemize}

The five EPF datasets are also among these 13 tasks. Their rolling-window probabilistic evaluation differs from the setting used for deterministic forecasting on the EPF datasets~\citep{wang2024timexer}.

\subsection{Electricity Price Forecasting Datasets}
The EPF collection~\citep{lago2021forecasting} comprises five hourly day-ahead electricity-price datasets, each spanning approximately six years. Their exogenous inputs are forecasts available before the prediction horizon, such as load or generation forecasts, rather than realized future observations.

NP (Nord Pool) spans 2013-01-01 to 2018-12-24 and includes electricity prices, grid load forecasts, and wind power forecasts. PJM (Pennsylvania-New Jersey-Maryland) covers the same period, with prices in the Commonwealth Edison (COMED) zone and forecasts of system and COMED load. BE (Belgium) spans 2011-01-09 to 2016-12-31 and includes electricity prices, Belgian load forecasts, and generation forecasts from neighboring France. FR (France) spans 2012-01-09 to 2017-12-31 and includes electricity prices alongside load and generation forecasts. DE (Germany) spans 2012-01-09 to 2017-12-31 and includes electricity prices, load forecasts for the TSO Amprion zone, and wind and solar generation forecasts.

\begin{table}[h]
    \centering
    \caption{Summary of the five Electricity Price Forecasting (EPF) datasets. The last column reports the numbers of sliding forecasting windows in the training, validation, and test splits, respectively, for $(W,H)=(168,24)$.}
    \label{tab:epf_summary}
    
    \setlength{\tabcolsep}{6pt}
    
    \begin{tabular}{l|c|l|c}
    \toprule
    \textbf{Dataset} & \textbf{Freq.} & \textbf{Target Variate} & \textbf{Forecast Windows (Train, Val, Test)} \\
    \midrule
    NP & Hourly & Electricity Price & (36500, 5219, 10460) \\
    PJM & Hourly & Electricity Price & (36500, 5219, 10460) \\
    BE & Hourly & Electricity Price & (36500, 5219, 10460) \\
    FR & Hourly & Electricity Price & (36500, 5219, 10460) \\
    DE & Hourly & Electricity Price & (36500, 5219, 10460) \\
    \bottomrule
    \end{tabular}
\end{table}

\begin{table}[h]
    \centering
    \caption{Summary of 13 probabilistic forecasting tasks selected from~\citet{shchur2025fev}. $H$ is the prediction horizon and $N_{\mathrm{roll}}$ is the number of rolling evaluation windows. Covariates are counted separately according to whether they are available only in the past or known over the forecast horizon.}
    \label{tab:fev_summary}
    \footnotesize
    \setlength{\tabcolsep}{3pt}    
    \begin{tabular}{l|c|c|c|r|c|c|c}
    \toprule
    \textbf{Dataset} & \textbf{Freq.} & \boldmath$H$ & \boldmath$N_{\mathrm{roll}}$ & \textbf{Median Length} & \textbf{\# Ser.} & \textbf{\# Past Cov.} & \textbf{\# Known Cov.} \\
    \midrule
    ENTSO-e Load & 15min  & 96 & 20 & 175,292 & 6 & 0 & 3 \\
    ENTSO-e Load & 30min  & 96 & 20 & 87,645 & 6 & 0 & 3 \\
    ENTSO-e Load & Hourly  & 168 & 20 & 43,822 & 6 & 0 & 3 \\
    EPF BE & Hourly & 24 & 20 & 52,416 & 1 & 0 & 2 \\
    EPF DE & Hourly & 24 & 20 & 52,416 & 1 & 0 & 2 \\
    EPF FR & Hourly & 24 & 20 & 52,416 & 1 & 0 & 2 \\
    EPF NP & Hourly & 24 & 20 & 52,416 & 1 & 0 & 2 \\
    EPF PJM & Hourly & 24 & 20 & 52,416 & 1 & 0 & 2 \\
    GFC12 & Hourly  & 168 & 10 & 39,414 & 11 & 0 & 1 \\
    GFC14 & Hourly  & 168 & 20 & 17,520 & 1 & 0 & 1 \\
    GFC17 & Hourly  & 168 & 20 & 17,544 & 8 & 0 & 1 \\
    Solar with Weather & 15min  & 96 & 20 & 198,600 & 1 & 2 & 7 \\
    Solar with Weather & Hourly  & 24 & 20 & 49,648 & 1 & 2 & 7 \\
    \bottomrule
    \end{tabular}
\end{table}

\section{Implementation Details}
\label{sec:implementation}

Our model is implemented using PyTorch~\citep{Paszke2019PyTorchAI}, and all experiments are conducted on a single NVIDIA RTX 4090 GPU (24GB).

\subsection{Flow Matching and Inference Configuration}
We sample training times from a logit-normal distribution, $t=\operatorname{sigmoid}(u)$ with $u\sim\mathcal{N}(0,1)$, and optimize the velocity loss in Equation~\ref{eq:velocity_loss}. Unless specified otherwise, inference uses five Euler integration steps per trajectory and averages ten independently initialized forecast samples for point prediction. The training-time schedule comparison is reported in Figure~\ref{fig:noise_schedule}; the sample-count and integration-step sensitivities are shown in Figure~\ref{fig:sampling_sensitivity}. Evaluation on the 13 selected probabilistic tasks uses 50 forecast samples, as specified below.

\subsection{Hyperparameter Settings}
We use separate hyperparameter grids for EPF and the 13 selected probabilistic tasks.

\paragraph{EPF Dataset.}
For EPF, we search the 12 combinations of learning rate in $\{10^{-4},3\times10^{-4},10^{-3}\}$, model dimension $d_{\text{model}}\in\{256,512\}$, and encoder depth $e_{\text{layers}}\in\{3,4\}$. Batch size is 32, patch length is 24, and the feed-forward dimension is $d_{\text{ff}}=4d_{\text{model}}$.

\paragraph{Selected Probabilistic Tasks.}
For each selected task, we evaluate DiTS under four $(d_{\mathrm{model}},d_{\mathrm{ff}},e_{\mathrm{layers}})$ configurations: $(256,768,3)$, $(256,1024,4)$, $(512,1536,3)$, and $(512,2048,4)$, each with learning rates in $\{10^{-4},3\times10^{-4},10^{-3}\}$. The patch length is 24 and the batch size is 64. Each baseline is evaluated over 12 hyperparameter configurations based on the settings reported in its original paper. For each task, all baselines use the same lookback length as DiTS. Table~\ref{tab:fev_hyperparams} reports the selected DiTS setting for all 13 tasks.

\begin{table}[ht]
    \centering
    \caption{Selected DiTS hyperparameters for the 13 probabilistic forecasting tasks. $W$ is the history length; the patch length is 24 and the batch size is 64 throughout.}
    \label{tab:fev_hyperparams}
    
    \footnotesize
    \setlength{\tabcolsep}{3pt} 
    \renewcommand{\arraystretch}{1.1} 
    
    \begin{tabular}{l|c|c|c|c|c|c|c}
        \toprule
        \textbf{Dataset} & \textbf{Freq.} & \boldmath{$W$} & \textbf{Batch} & \boldmath{$d_{\text{model}}$} & \boldmath{$d_{\text{ff}}$} & \textbf{LR} & \boldmath{$e_{\text{layers}}$} \\
        \midrule
        ENTSO-e Load & 15min & 720 & 64 & 512 & 2048 & $10^{-3}$ & 4 \\
        ENTSO-e Load & 30min & 720 & 64 & 512 & 2048 & $10^{-4}$ & 4 \\
        ENTSO-e Load & Hourly & 720 & 64 & 512 & 2048 & $10^{-3}$ & 4 \\
        GFC12 & Hourly & 720 & 64 & 256 & 768 & $3\times10^{-4}$ & 3 \\
        GFC14 & Hourly & 720 & 64 & 256 & 1024 & $3\times10^{-4}$ & 4 \\
        GFC17 & Hourly & 720 & 64 & 256 & 768 & $10^{-3}$ & 3 \\
        Solar with Weather & 15min & 720 & 64 & 512 & 1536 & $10^{-4}$ & 3 \\
        Solar with Weather & Hourly & 336 & 64 & 256 & 768 & $10^{-3}$ & 3 \\
        EPF BE & Hourly & 720 & 64 & 512 & 1536 & $10^{-3}$ & 3 \\
        EPF DE & Hourly & 720 & 64 & 512 & 2048 & $10^{-4}$ & 4 \\
        EPF FR & Hourly & 720 & 64 & 256 & 1024 & $3\times10^{-4}$ & 4 \\
        EPF NP & Hourly & 720 & 64 & 512 & 2048 & $10^{-3}$ & 4 \\
        EPF PJM & Hourly & 720 & 64 & 256 & 768 & $10^{-4}$ & 3 \\
        \bottomrule
    \end{tabular}
\end{table}

\subsection{Protocols for Additional Experiments}
\label{sec:additional_protocols}
We report additional ablations on prediction parameterization, generative noise paths, numerical solvers, classifier-free guidance, and hyperparameter sensitivity in Appendices~\ref{sec:prediction_objectives}--\ref{sec:hyper_sensitivity}. Ablations requiring retraining use the 12-configuration EPF grid described above, and the reported results are selected from this grid separately for each dataset and experimental setting. Table~\ref{tab:flow_covariate_ablation} separately compares covariate-availability settings.

\section{Additional Analysis}

\subsection{Prediction and Loss Parameterizations}
\label{sec:prediction_objectives}
In a flow-matching formulation, \citet{li2025backtobasics} separate the network's prediction space from the loss space and advocate directly predicting clean data $x$ rather than velocity $v$, while retaining a flow-matching objective in velocity space. Motivated by this distinction, we evaluate the four combinations of $x$- and $v$-prediction with $x$- and $v$-loss for time-series forecasting. Table~\ref{tab:prediction_objectives} reports the corresponding EPF results.

\begin{table}[htbp]
\centering
\caption{Prediction and loss parameterizations on five EPF datasets with known-future covariates and $(W,H)=(168,24)$. Each column pair specifies the model's direct output ($x$ or $v$) and the space in which its loss is computed. MSE and MAE are lower-is-better; red boldface marks the best unrounded value for each dataset and metric.}
\label{tab:prediction_objectives}
\small
\setlength{\tabcolsep}{6pt}
\renewcommand{\arraystretch}{1.1}
\begin{tabular}{lcccccccc}
\toprule
& \multicolumn{2}{c}{$v$-pred/$v$-loss} & \multicolumn{2}{c}{$x$-pred/$x$-loss} & \multicolumn{2}{c}{$v$-pred/$x$-loss} & \multicolumn{2}{c}{$x$-pred/$v$-loss} \\
\cmidrule(lr){2-3}\cmidrule(lr){4-5}\cmidrule(lr){6-7}\cmidrule(lr){8-9}
Dataset & MSE & MAE & MSE & MAE & MSE & MAE & MSE & MAE \\
\midrule
NP  & \textcolor{red}{\textbf{0.160}} & \textcolor{red}{\textbf{0.205}} & 0.164 & 0.209 & 0.161 & 0.211 & 0.170 & 0.213 \\
PJM & 0.056 & \textcolor{red}{\textbf{0.141}} & 0.057 & 0.142 & 0.057 & 0.143 & \textcolor{red}{\textbf{0.056}} & 0.142 \\
BE  & 0.325 & \textcolor{red}{\textbf{0.232}} & 0.325 & 0.235 & \textcolor{red}{\textbf{0.319}} & 0.234 & 0.361 & 0.252 \\
FR  & \textcolor{red}{\textbf{0.269}} & \textcolor{red}{\textbf{0.171}} & 0.282 & 0.172 & 0.282 & 0.174 & 0.289 & 0.179 \\
DE  & \textcolor{red}{\textbf{0.208}} & \textcolor{red}{\textbf{0.286}} & 0.216 & 0.292 & 0.216 & 0.294 & 0.230 & 0.300 \\
\midrule
AVG & \textcolor{red}{\textbf{0.204}} & \textcolor{red}{\textbf{0.207}} & 0.209 & 0.210 & 0.207 & 0.211 & 0.221 & 0.217 \\
\bottomrule
\end{tabular}
\end{table}

The $v$-prediction/$v$-loss configuration yields the lowest five-dataset average MSE and MAE among the four evaluated parameterizations. This differs from the preference for $x$-prediction on high-dimensional image patches reported by \citet{li2025backtobasics}, suggesting that the preferred parameterization may depend on the data modality.

\subsection{Generative Noise Paths}
\label{sec:diffusion_frameworks}
We compare rectified flow, DDPM, and EDM as alternative generative noise paths. Table~\ref{tab:diffusion_frameworks} reports forecasting accuracy for each EPF dataset and on average.

\begin{table}[H]
\centering
\caption{Generative noise-path comparison on five EPF datasets with known-future covariates and $(W,H)=(168,24)$. MSE and MAE are lower-is-better; red boldface marks the best unrounded value for each dataset and metric.}
\label{tab:diffusion_frameworks}
\small
\setlength{\tabcolsep}{6pt}
\renewcommand{\arraystretch}{1.1}
\begin{tabular}{lcccccc}
\toprule
& \multicolumn{2}{c}{Rectified flow} & \multicolumn{2}{c}{DDPM} & \multicolumn{2}{c}{EDM} \\
\cmidrule(lr){2-3}\cmidrule(lr){4-5}\cmidrule(lr){6-7}
Dataset & MSE & MAE & MSE & MAE & MSE & MAE \\
\midrule
NP  & \textcolor{red}{\textbf{0.160}} & \textcolor{red}{\textbf{0.205}} & 0.171 & 0.216 & 0.221 & 0.268 \\
PJM & \textcolor{red}{\textbf{0.056}} & \textcolor{red}{\textbf{0.141}} & 0.059 & 0.144 & 0.076 & 0.178 \\
BE  & \textcolor{red}{\textbf{0.325}} & 0.232 & 0.326 & \textcolor{red}{\textbf{0.224}} & 0.431 & 0.306 \\
FR  & \textcolor{red}{\textbf{0.269}} & 0.171 & 0.277 & \textcolor{red}{\textbf{0.167}} & 0.351 & 0.226 \\
DE  & \textcolor{red}{\textbf{0.208}} & \textcolor{red}{\textbf{0.286}} & 0.243 & 0.313 & 0.323 & 0.377 \\
\midrule
AVG & \textcolor{red}{\textbf{0.204}} & \textcolor{red}{\textbf{0.207}} & 0.215 & 0.213 & 0.280 & 0.271 \\
\bottomrule
\end{tabular}
\end{table}

Among the evaluated generative noise paths, rectified flow yields the lowest five-dataset average MSE and MAE under this setup. This observation is consistent with reports from image and video generation~\citep{esser2024scaling,chen2025goku} and with the motivation for straighter transport paths~\citep{liu2022flow}.

\subsection{Numerical Solvers}
\label{sec:numerical_solvers}

We compare Euler, Heun, and fourth-order Runge--Kutta (RK4) to assess whether a higher-order solver improves forecasting. Table~\ref{tab:numerical_solvers} reports their EPF results.

\begin{table}[H]
\centering
\caption{Numerical-solver ablation on five EPF datasets with known-future covariates and $(W,H)=(168,24)$. MSE and MAE are lower-is-better; red boldface marks the best unrounded value for each dataset and metric.}
\label{tab:numerical_solvers}
\small
\setlength{\tabcolsep}{7pt}
\renewcommand{\arraystretch}{1.1}
\begin{tabular}{lcccccc}
\toprule
& \multicolumn{2}{c}{Euler} & \multicolumn{2}{c}{Heun} & \multicolumn{2}{c}{RK4} \\
\cmidrule(lr){2-3}\cmidrule(lr){4-5}\cmidrule(lr){6-7}
Dataset & MSE & MAE & MSE & MAE & MSE & MAE \\
\midrule
NP  & \textcolor{red}{\textbf{0.160}} & \textcolor{red}{\textbf{0.205}} & 0.167 & 0.209 & 0.169 & 0.211 \\
PJM & \textcolor{red}{\textbf{0.056}} & \textcolor{red}{\textbf{0.141}} & 0.058 & 0.145 & 0.059 & 0.144 \\
BE  & \textcolor{red}{\textbf{0.325}} & \textcolor{red}{\textbf{0.232}} & 0.326 & 0.235 & 0.329 & 0.236 \\
FR  & 0.269 & \textcolor{red}{\textbf{0.171}} & \textcolor{red}{\textbf{0.268}} & 0.174 & 0.269 & 0.174 \\
DE  & \textcolor{red}{\textbf{0.208}} & \textcolor{red}{\textbf{0.286}} & 0.213 & 0.290 & 0.214 & 0.290 \\
\midrule
AVG & \textcolor{red}{\textbf{0.204}} & \textcolor{red}{\textbf{0.207}} & 0.206 & 0.211 & 0.208 & 0.211 \\
\bottomrule
\end{tabular}
\end{table}

Euler achieves the lowest five-dataset average MSE and MAE, and leads on every dataset-level metric except FR MSE, where Heun is slightly lower (0.268 versus 0.269). Heun and RK4 require more model forward passes without improving average forecasting accuracy, so these results favor Euler under the evaluated inference budget.

\subsection{Classifier-Free Guidance}
\label{sec:cfg_guidance}
\begin{figure}[H]
\centering
\includegraphics[width=\textwidth]{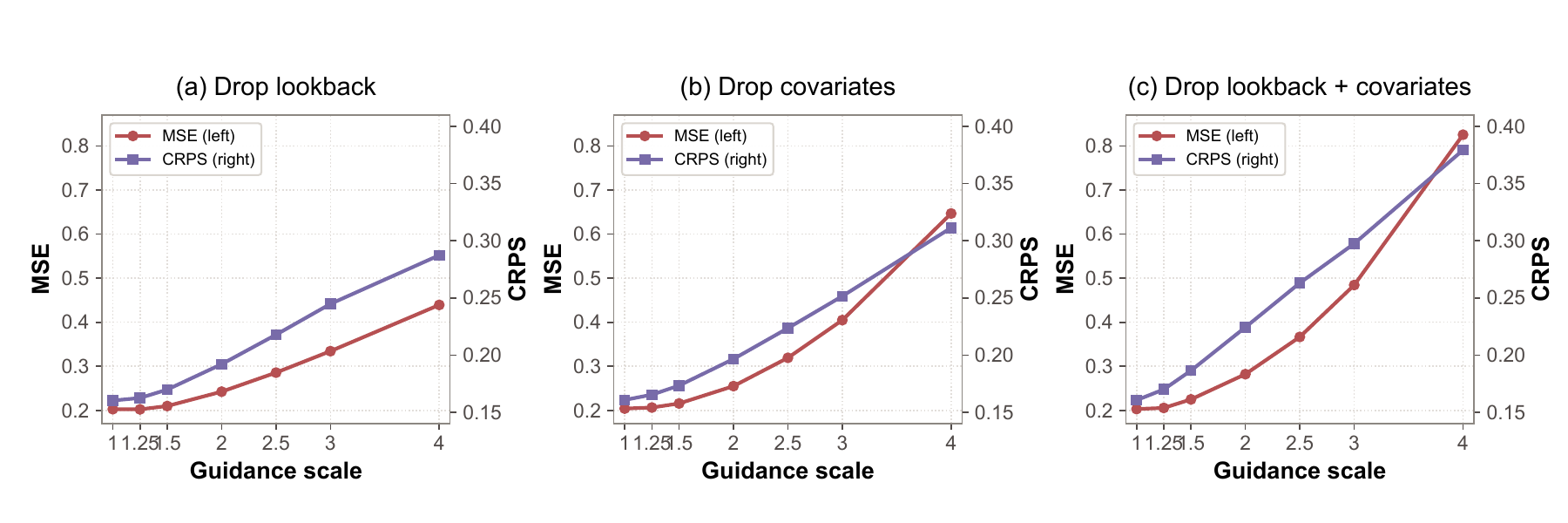}
\caption{Classifier-free guidance sensitivity on five EPF datasets with known-future covariates and $(W,H)=(168,24)$. Panels (a)--(c) drop the target lookback, covariates, or both in the guidance branch, respectively. Curves show the arithmetic mean MSE (left axis) and CRPS (right axis) across NP, PJM, BE, FR, and DE; lower is better.}
\label{fig:cfg_guidance}
\end{figure}

We vary the guidance scale and the conditions removed in the guidance branch: target lookback only, covariates only, or both. For conditional velocity $\mathbf{v}_{\mathrm{cond}}$ and the corresponding null-condition velocity $\mathbf{v}_{\mathrm{null}}$ evaluated at the same noisy future state and diffusion time, guidance scale $s$ gives
\begin{equation}
\mathbf{v}_{\mathrm{CFG}}=\mathbf{v}_{\mathrm{null}}+s\left(\mathbf{v}_{\mathrm{cond}}-\mathbf{v}_{\mathrm{null}}\right).
\end{equation}
In all three variants, each training example drops the specified condition with probability 0.1. The dropped target-history or covariate condition is replaced by zeros after encoding; when both are dropped, one shared Bernoulli decision zeros both representations. The noisy future state and diffusion time are always retained; any condition not selected for dropout is also retained. Figure~\ref{fig:cfg_guidance} plots the five-dataset average MSE and CRPS for each choice under the default EPF ablation protocol.

Across all three condition-drop settings, guidance scale 1, which does not amplify the conditional--unconditional difference, gives the most consistent results. Increasing the scale generally degrades both point accuracy and probabilistic forecast quality, especially when both conditions are dropped.

\subsection{Hyperparameter Sensitivity}
\label{sec:hyper_sensitivity}
Table~\ref{tab:hyper_grid} reports the complete EPF hyperparameter grid described in Section~\ref{sec:implementation}, allowing sensitivity to learning rate, model width, and depth to be assessed across datasets.

\begin{table}[htbp]
\centering
\caption{Complete 12-configuration EPF grid with known-future covariates and $(W,H)=(168,24)$. LR is learning rate, $d=d_{\mathrm{model}}$, $e=e_{\mathrm{layers}}$, and $d_{\mathrm{ff}}=4d$ throughout. Entries are MSE (lower is better). Red boldface marks the reported best unrounded configuration in each column; values may tie after rounding.}
\label{tab:hyper_grid}
\footnotesize
\setlength{\tabcolsep}{4pt}
\renewcommand{\arraystretch}{1.1}
\begin{tabular}{lcccccccc}
\toprule
LR & $d$ & $e$ & NP & PJM & BE & FR & DE & AVG \\
\midrule
$10^{-4}$ & 256 & 3 & 0.169 & 0.057 & 0.330 & 0.287 & 0.248 & 0.218 \\
$3 \times 10^{-4}$ & 256 & 4 & 0.166 & 0.056 & 0.333 & 0.280 & 0.218 & 0.211 \\
$10^{-3}$ & 256 & 3 & 0.165 & 0.056 & 0.338 & 0.274 & 0.220 & 0.211 \\
$10^{-4}$ & 256 & 4 & 0.168 & 0.058 & 0.333 & 0.292 & 0.244 & 0.219 \\
$3 \times 10^{-4}$ & 256 & 3 & 0.168 & 0.056 & 0.332 & 0.287 & 0.218 & 0.212 \\
$10^{-3}$ & 256 & 4 & 0.160 & 0.057 & 0.335 & \textcolor{red}{\textbf{0.269}} & \textcolor{red}{\textbf{0.208}} & \textcolor{red}{\textbf{0.206}} \\
$10^{-4}$ & 512 & 3 & 0.165 & 0.056 & \textcolor{red}{\textbf{0.325}} & 0.286 & 0.235 & 0.213 \\
$3 \times 10^{-4}$ & 512 & 4 & 0.161 & 0.058 & 0.325 & 0.277 & 0.212 & 0.207 \\
$10^{-3}$ & 512 & 3 & 0.249 & 0.060 & 0.349 & 0.344 & 0.216 & 0.244 \\
$10^{-4}$ & 512 & 4 & 0.165 & \textcolor{red}{\textbf{0.056}} & 0.330 & 0.289 & 0.222 & 0.212 \\
$3 \times 10^{-4}$ & 512 & 3 & \textcolor{red}{\textbf{0.160}} & 0.057 & 0.333 & 0.284 & 0.217 & 0.210 \\
$10^{-3}$ & 512 & 4 & 0.225 & 0.060 & 0.334 & 0.336 & 0.215 & 0.234 \\
\bottomrule
\end{tabular}
\end{table}

\clearpage
\section{Full Results}
\label{sec:full_results}

\subsection{Flow Matching and Covariate Availability}
Table~\ref{tab:flow_covariate_ablation} reports the per-dataset MSE for flow matching and direct MSE regression under three covariate-availability settings, complementing the aggregate comparison in Figure~\ref{fig:flow_covariate_mse}.

\begin{table}[H]
\centering
\caption{Flow matching and direct MSE regression on five EPF datasets with $(W,H)=(168,24)$. The three covariate-availability settings are known-future covariates, past-only covariates, and target history only. The latter two settings are exceptions to the default known-future ablation protocol. Entries are MSE (lower is better); red boldface marks the better unrounded value within each method pair.}
\label{tab:flow_covariate_ablation}
\small
\setlength{\tabcolsep}{5pt}
\begin{tabular}{l|cc|cc|cc}
\toprule
 & \multicolumn{2}{c|}{Known-future covariates} & \multicolumn{2}{c|}{Past-only covariates} & \multicolumn{2}{c}{Target only} \\
\cmidrule(lr){2-3}\cmidrule(lr){4-5}\cmidrule(lr){6-7}
Dataset & Flow Matching & Direct MSE & Flow Matching & Direct MSE & Flow Matching & Direct MSE \\
\midrule
NP  & \textcolor{red}{\textbf{0.160}} & 0.179 & \textcolor{red}{\textbf{0.218}} & 0.241 & 0.246 & \textcolor{red}{\textbf{0.240}} \\
PJM & \textcolor{red}{\textbf{0.056}} & 0.070 & \textcolor{red}{\textbf{0.069}} & 0.075 & \textcolor{red}{\textbf{0.077}} & 0.081 \\
BE  & \textcolor{red}{\textbf{0.325}} & 0.326 & \textcolor{red}{\textbf{0.370}} & 0.376 & 0.393 & \textcolor{red}{\textbf{0.373}} \\
FR  & \textcolor{red}{\textbf{0.269}} & 0.320 & \textcolor{red}{\textbf{0.341}} & 0.386 & \textcolor{red}{\textbf{0.362}} & 0.378 \\
DE  & \textcolor{red}{\textbf{0.208}} & 0.242 & \textcolor{red}{\textbf{0.403}} & 0.434 & 0.455 & \textcolor{red}{\textbf{0.436}} \\
\midrule
AVG & \textcolor{red}{\textbf{0.204}} & 0.228 & \textcolor{red}{\textbf{0.280}} & 0.302 & 0.307 & \textcolor{red}{\textbf{0.302}} \\
\bottomrule
\end{tabular}
\end{table}

\subsection{Forecasting with Exogenous Covariates}
\label{subsec:exogenous_results}

We report full deterministic results for EPF~\citep{wang2024timexer} and probabilistic results for the 13 tasks selected from~\citet{shchur2025fev}.
\paragraph{Experimental Setup and Protocols.}

For \textbf{EPF}, we evaluate $(W,H)=(168,24)$ both with known-future covariates and with covariates available only in the historical window. We also report a longer EPF prediction horizon, $(W,H)=(720,360)$, with known-future covariates. Table~\ref{tab:epf_wi_future_full} gives the full known-future results; Table~\ref{tab:epf_wo_future} in the main paper gives the $H=24$ results without future covariates.

For the \textbf{13 selected probabilistic tasks}, Table~\ref{tab:fev_full} compares DiTS with DeepAR, TimeGrad, TempFlow, and GCGNet using known-future covariates.

We evaluate DiTS under the 12 configurations described in Section~\ref{sec:implementation}; each baseline is likewise evaluated over 12 configurations per task based on its published settings. For each task, the validation set consists of the windows immediately preceding the test period and matches the test set in window count. To limit computation, we retain at most 100,000 training windows per task, selecting those closest in time to the validation period. We exclude past-only covariates from training inputs for all models because some baselines do not support them.

Following~\citet{shchur2025fev}, we report Scaled Quantile Loss (SQL), Mean Absolute Scaled Error (MASE), Weighted Absolute Percentage Error (WAPE), and Weighted Quantile Loss (WQL). SQL and MASE normalize errors by each series' historical seasonal error, whereas WQL and WAPE aggregate errors in a scale-dependent manner. We report each metric for every task and its geometric mean across the 13 tasks.

\begingroup
\scriptsize
\setlength{\tabcolsep}{5pt}
\renewcommand{\arraystretch}{1.08}
\begin{longtable}{lc|ccccc}
\caption{Probabilistic forecasting on 13 selected tasks with known-future covariates. SQL, MASE, WAPE, and WQL are lower-is-better metrics. Red boldface and blue underlining mark the best and second-best unrounded values in each row, respectively. The geometric mean is calculated separately for each metric over the 13 tasks.}\label{tab:fev_full}\\
\toprule
\multirow{2}{*}{Dataset} & \multirow{2}{*}{Metric} & \multicolumn{5}{c}{Models} \\
\cmidrule(lr){3-7}
 & & DiTS & DeepAR & TimeGrad & TempFlow & GCGNet \\
\midrule
\endfirsthead
\caption[]{Probabilistic forecasting on 13 selected tasks (continued).}\\
\toprule
\multirow{2}{*}{Dataset} & \multirow{2}{*}{Metric} & \multicolumn{5}{c}{Models} \\
\cmidrule(lr){3-7}
 & & DiTS & DeepAR & TimeGrad & TempFlow & GCGNet \\
\midrule
\endhead
\bottomrule
\endfoot
\multirow{4}{*}{ENTSO-e Load (15T)} & SQL & 0.829 & \textcolor{blue}{\underline{0.801}} & 0.957 & 0.936 & \textcolor{red}{\textbf{0.779}} \\
 & MASE & 0.968 & \textcolor{blue}{\underline{0.945}} & 1.086 & 1.128 & \textcolor{red}{\textbf{0.779}} \\
 & WAPE & 0.077 & \textcolor{blue}{\underline{0.067}} & 0.079 & 0.078 & \textcolor{red}{\textbf{0.057}} \\
 & WQL & 0.067 & \textcolor{blue}{\underline{0.058}} & 0.070 & 0.065 & \textcolor{red}{\textbf{0.057}} \\
\midrule
\multirow{4}{*}{ENTSO-e Load (30T)} & SQL & \textcolor{blue}{\underline{0.577}} & \textcolor{red}{\textbf{0.559}} & 0.668 & 0.871 & 0.594 \\*
 & MASE & \textcolor{blue}{\underline{0.681}} & 0.684 & 0.809 & 1.069 & \textcolor{red}{\textbf{0.597}} \\*
 & WAPE & 0.055 & \textcolor{blue}{\underline{0.046}} & 0.057 & 0.077 & \textcolor{red}{\textbf{0.041}} \\*
 & WQL & 0.047 & \textcolor{red}{\textbf{0.039}} & 0.047 & 0.063 & \textcolor{blue}{\underline{0.041}} \\
\midrule
\multirow{4}{*}{ENTSO-e Load (1H)} & SQL & \textcolor{red}{\textbf{0.480}} & 0.528 & 0.802 & \textcolor{blue}{\underline{0.525}} & 0.650 \\
 & MASE & \textcolor{red}{\textbf{0.555}} & \textcolor{blue}{\underline{0.641}} & 0.985 & 0.645 & 0.652 \\
 & WAPE & \textcolor{red}{\textbf{0.037}} & \textcolor{blue}{\underline{0.041}} & 0.064 & 0.042 & 0.043 \\
 & WQL & \textcolor{red}{\textbf{0.032}} & \textcolor{blue}{\underline{0.034}} & 0.052 & 0.034 & 0.043 \\
\midrule
\multirow{4}{*}{GFC12} & SQL & \textcolor{red}{\textbf{0.721}} & \textcolor{blue}{\underline{0.828}} & 1.163 & 0.978 & 0.940 \\
 & MASE & \textcolor{red}{\textbf{0.881}} & \textcolor{blue}{\underline{1.015}} & 1.447 & 1.207 & 1.046 \\
 & WAPE & \textcolor{red}{\textbf{0.073}} & \textcolor{blue}{\underline{0.089}} & 0.129 & 0.108 & 0.092 \\
 & WQL & \textcolor{red}{\textbf{0.059}} & \textcolor{blue}{\underline{0.073}} & 0.102 & 0.086 & 0.083 \\
\midrule
\multirow{4}{*}{GFC14} & SQL & \textcolor{red}{\textbf{0.471}} & \textcolor{blue}{\underline{0.601}} & 0.683 & 0.635 & 1.079 \\
 & MASE & \textcolor{red}{\textbf{0.564}} & \textcolor{blue}{\underline{0.749}} & 0.819 & 0.793 & 1.081 \\
 & WAPE & \textcolor{red}{\textbf{0.028}} & \textcolor{blue}{\underline{0.037}} & 0.040 & 0.039 & 0.053 \\
 & WQL & \textcolor{red}{\textbf{0.023}} & \textcolor{blue}{\underline{0.030}} & 0.034 & 0.031 & 0.053 \\
\midrule
\multirow{4}{*}{GFC17} & SQL & \textcolor{red}{\textbf{0.498}} & \textcolor{blue}{\underline{0.769}} & 0.900 & 0.940 & 1.032 \\
 & MASE & \textcolor{red}{\textbf{0.598}} & \textcolor{blue}{\underline{0.942}} & 1.074 & 1.142 & 1.110 \\
 & WAPE & \textcolor{red}{\textbf{0.039}} & \textcolor{blue}{\underline{0.063}} & 0.069 & 0.075 & 0.074 \\
 & WQL & \textcolor{red}{\textbf{0.032}} & \textcolor{blue}{\underline{0.052}} & 0.058 & 0.061 & 0.069 \\
\midrule
\multirow{4}{*}{Solar with Weather (15T)} & SQL & \textcolor{blue}{\underline{0.729}} & \textcolor{red}{\textbf{0.710}} & 0.865 & 0.873 & 0.960 \\*
 & MASE & \textcolor{blue}{\underline{0.891}} & \textcolor{red}{\textbf{0.871}} & 1.038 & 1.050 & 0.970 \\*
 & WAPE & \textcolor{blue}{\underline{0.995}} & \textcolor{red}{\textbf{0.986}} & 1.219 & 1.353 & 1.079 \\*
 & WQL & \textcolor{blue}{\underline{0.839}} & \textcolor{red}{\textbf{0.808}} & 1.017 & 1.123 & 1.071 \\
\midrule
\multirow{4}{*}{Solar with Weather (1H)} & SQL & \textcolor{red}{\textbf{0.691}} & \textcolor{blue}{\underline{0.841}} & 0.854 & 0.888 & 0.996 \\
 & MASE & \textcolor{red}{\textbf{0.859}} & 1.021 & 1.030 & 1.066 & \textcolor{blue}{\underline{1.019}} \\
 & WAPE & \textcolor{red}{\textbf{0.909}} & 1.139 & \textcolor{blue}{\underline{1.036}} & 1.146 & 1.098 \\
 & WQL & \textcolor{red}{\textbf{0.782}} & 0.963 & \textcolor{blue}{\underline{0.855}} & 0.953 & 1.080 \\
\midrule
\multirow{4}{*}{EPF BE} & SQL & \textcolor{red}{\textbf{0.543}} & 0.581 & 0.551 & \textcolor{blue}{\underline{0.543}} & 0.733 \\
 & MASE & \textcolor{red}{\textbf{0.682}} & 0.727 & \textcolor{blue}{\underline{0.695}} & 0.697 & 0.735 \\
 & WAPE & \textcolor{red}{\textbf{0.123}} & 0.132 & 0.124 & \textcolor{blue}{\underline{0.124}} & 0.134 \\
 & WQL & \textcolor{blue}{\underline{0.098}} & 0.105 & 0.098 & \textcolor{red}{\textbf{0.096}} & 0.133 \\
\midrule
\multirow{4}{*}{EPF DE} & SQL & \textcolor{red}{\textbf{0.497}} & 0.667 & 0.554 & \textcolor{blue}{\underline{0.529}} & 0.627 \\
 & MASE & \textcolor{red}{\textbf{0.616}} & 0.820 & 0.679 & 0.684 & \textcolor{blue}{\underline{0.631}} \\
 & WAPE & \textcolor{red}{\textbf{0.264}} & 0.326 & 0.437 & 0.311 & \textcolor{blue}{\underline{0.294}} \\
 & WQL & \textcolor{red}{\textbf{0.213}} & 0.273 & 0.348 & \textcolor{blue}{\underline{0.243}} & 0.292 \\
\midrule
\multirow{4}{*}{EPF FR} & SQL & \textcolor{red}{\textbf{0.358}} & 0.425 & \textcolor{blue}{\underline{0.365}} & 0.394 & 0.509 \\
 & MASE & \textcolor{red}{\textbf{0.458}} & 0.539 & \textcolor{blue}{\underline{0.463}} & 0.495 & 0.511 \\
 & WAPE & \textcolor{red}{\textbf{0.068}} & 0.082 & \textcolor{blue}{\underline{0.069}} & 0.074 & 0.076 \\
 & WQL & \textcolor{red}{\textbf{0.054}} & 0.064 & \textcolor{blue}{\underline{0.054}} & 0.058 & 0.076 \\
\midrule
\multirow{4}{*}{EPF NP} & SQL & 0.685 & 0.949 & \textcolor{red}{\textbf{0.624}} & \textcolor{blue}{\underline{0.659}} & 0.916 \\
 & MASE & \textcolor{blue}{\underline{0.835}} & 1.140 & \textcolor{red}{\textbf{0.797}} & 0.855 & 0.919 \\
 & WAPE & \textcolor{blue}{\underline{0.038}} & 0.051 & \textcolor{red}{\textbf{0.036}} & 0.039 & 0.041 \\
 & WQL & 0.031 & 0.042 & \textcolor{red}{\textbf{0.028}} & \textcolor{blue}{\underline{0.030}} & 0.041 \\
\midrule
\multirow{4}{*}{EPF PJM} & SQL & 0.386 & \textcolor{red}{\textbf{0.342}} & \textcolor{blue}{\underline{0.374}} & 0.385 & 0.458 \\
 & MASE & 0.479 & \textcolor{red}{\textbf{0.420}} & 0.465 & 0.484 & \textcolor{blue}{\underline{0.459}} \\
 & WAPE & 0.084 & \textcolor{red}{\textbf{0.075}} & 0.082 & 0.085 & \textcolor{blue}{\underline{0.080}} \\
 & WQL & 0.068 & \textcolor{red}{\textbf{0.061}} & \textcolor{blue}{\underline{0.066}} & 0.068 & 0.079 \\
\midrule
\multirow{4}{*}{\textbf{Geometric Mean}} & SQL & \textcolor{red}{\textbf{0.557}} & \textcolor{blue}{\underline{0.638}} & 0.683 & 0.671 & 0.763 \\
 & MASE & \textcolor{red}{\textbf{0.678}} & 0.782 & 0.835 & 0.833 & \textcolor{blue}{\underline{0.777}} \\
 & WAPE & \textcolor{red}{\textbf{0.098}} & \textcolor{blue}{\underline{0.111}} & 0.123 & 0.121 & 0.112 \\
 & WQL & \textcolor{red}{\textbf{0.081}} & \textcolor{blue}{\underline{0.091}} & 0.100 & 0.097 & 0.110 \\
\end{longtable}
\endgroup

\begin{table}[H]
\centering
\caption{Full EPF forecasting results with known-future covariates for $(W,H)=(168,24)$ and $(720,360)$. The baselines follow Table~\ref{tab:epf_wi_future}; GCGNet and DAG results are from~\citet{li2026gcgnet} and~\citet{qiu2025dag}, respectively. Lower MSE and MAE values are better. AVG matches the two-horizon summary in Table~\ref{tab:epf_wi_future}. Red boldface and blue underlining mark the best and second-best displayed values in each row and metric.}
\label{tab:epf_wi_future_full}
\small
\setlength{\tabcolsep}{3.5pt}
\resizebox{\textwidth}{!}{\begin{tabular}{cc|cc|cc|cc|cc|cc|cc|cc|cc}
\toprule
\multicolumn{2}{c|}{Models} & \multicolumn{2}{c|}{DiTS} & \multicolumn{2}{c|}{GCGNet} & \multicolumn{2}{c|}{TimeXer} & \multicolumn{2}{c|}{DAG} & \multicolumn{2}{c|}{iTrans.} & \multicolumn{2}{c|}{Cross.} & \multicolumn{2}{c|}{DUET} & \multicolumn{2}{c}{TFT} \\
\multicolumn{2}{c|}{Metrics} & MSE & MAE & MSE & MAE & MSE & MAE & MSE & MAE & MSE & MAE & MSE & MAE & MSE & MAE & MSE & MAE \\ \midrule
\multirow{3}{*}{\rotatebox{90}{NP}}
 & 24 & \boldres{0.160} & \boldres{0.205} & 0.197 & 0.233 & \secondres{0.181} & \secondres{0.229} & 0.202 & 0.237 & 0.192 & 0.250 & 0.196 & 0.241 & 0.246 & 0.287 & 0.219 & 0.249 \\
 & 360 & \secondres{0.421} & \secondres{0.426} & 0.496 & 0.442 & 0.568 & 0.482 & 0.521 & 0.451 & 0.516 & 0.496 & \boldres{0.404} & \boldres{0.406} & 0.576 & 0.528 & 0.539 & 0.501 \\ \cmidrule{2-18}
 & AVG & \boldres{0.290} & \boldres{0.315} & 0.346 & 0.337 & 0.375 & 0.355 & 0.362 & 0.344 & 0.354 & 0.373 & \secondres{0.300} & \secondres{0.323} & 0.411 & 0.408 & 0.379 & 0.375 \\ \midrule
\multirow{3}{*}{\rotatebox{90}{PJM}}
 & 24 & \boldres{0.056} & \boldres{0.141} & 0.058 & 0.148 & 0.079 & 0.175 & \secondres{0.057} & \secondres{0.143} & 0.075 & 0.164 & 0.087 & 0.188 & 0.072 & 0.166 & 0.095 & 0.195 \\
 & 360 & \boldres{0.114} & \boldres{0.201} & 0.128 & 0.223 & 0.170 & 0.275 & 0.130 & \secondres{0.218} & \secondres{0.120} & 0.231 & 0.152 & 0.236 & 0.131 & 0.228 & 0.133 & 0.219 \\ \cmidrule{2-18}
 & AVG & \boldres{0.085} & \boldres{0.171} & \secondres{0.093} & 0.186 & 0.124 & 0.225 & \secondres{0.093} & \secondres{0.180} & 0.098 & 0.197 & 0.119 & 0.212 & 0.102 & 0.197 & 0.114 & 0.207 \\ \midrule
\multirow{3}{*}{\rotatebox{90}{BE}}
 & 24 & \secondres{0.325} & 0.232 & \boldres{0.323} & \boldres{0.227} & 0.377 & 0.237 & 0.361 & 0.229 & 0.343 & 0.250 & 0.345 & \secondres{0.228} & 0.432 & 0.272 & 0.426 & 0.272 \\
 & 360 & \boldres{0.425} & \boldres{0.291} & 0.524 & 0.347 & 0.486 & 0.322 & 0.485 & 0.330 & 0.684 & 0.441 & \secondres{0.461} & \secondres{0.308} & 0.597 & 0.436 & 0.482 & 0.310 \\ \cmidrule{2-18}
 & AVG & \boldres{0.375} & \boldres{0.261} & 0.423 & 0.287 & 0.432 & 0.279 & 0.423 & 0.279 & 0.513 & 0.345 & \secondres{0.403} & \secondres{0.268} & 0.515 & 0.354 & 0.454 & 0.291 \\ \midrule
\multirow{3}{*}{\rotatebox{90}{FR}}
 & 24 & \boldres{0.269} & \boldres{0.171} & \secondres{0.332} & \secondres{0.179} & 0.359 & 0.193 & 0.355 & \boldres{0.171} & 0.369 & 0.203 & 0.367 & 0.194 & 0.384 & 0.251 & 0.543 & 0.253 \\
 & 360 & \boldres{0.394} & \boldres{0.234} & 0.478 & 0.280 & 0.471 & 0.267 & 0.473 & 0.268 & 0.599 & 0.319 & \secondres{0.457} & 0.270 & 0.607 & 0.403 & 0.465 & \secondres{0.261} \\ \cmidrule{2-18}
 & AVG & \boldres{0.332} & \boldres{0.202} & \secondres{0.405} & 0.230 & 0.415 & 0.230 & 0.414 & \secondres{0.219} & 0.484 & 0.261 & 0.412 & 0.232 & 0.496 & 0.327 & 0.504 & 0.257 \\ \midrule
\multirow{3}{*}{\rotatebox{90}{DE}}
 & 24 & \boldres{0.208} & \boldres{0.286} & 0.282 & 0.328 & 0.287 & 0.328 & 0.277 & \secondres{0.322} & \secondres{0.261} & 0.325 & 0.268 & 0.323 & 0.376 & 0.378 & 0.380 & 0.383 \\
 & 360 & \boldres{0.331} & \boldres{0.348} & 0.491 & 0.445 & 0.526 & 0.453 & 0.462 & \secondres{0.418} & 0.555 & 0.488 & \secondres{0.444} & 0.430 & 0.589 & 0.482 & 0.599 & 0.509 \\ \cmidrule{2-18}
 & AVG & \boldres{0.270} & \boldres{0.317} & 0.387 & 0.387 & 0.406 & 0.391 & 0.370 & \secondres{0.370} & 0.408 & 0.407 & \secondres{0.356} & 0.376 & 0.482 & 0.430 & 0.489 & 0.446 \\ \bottomrule
\end{tabular}}
\end{table}

\subsection{Covariate Conditioning Mechanisms}
\label{sec:covariate_conditioning_appendix}
Table~\ref{tab:ablation_cond_appendix} gives the full results for Time-aligned Modulation in DiTS and four alternative conditioning mechanisms: cross-attention, global modulation, joint attention, and variate attention. Figure~\ref{fig:covariate_conditioning} illustrates these mechanisms.

The four alternatives differ in how covariates enter the target computation. \emph{Cross-attention} lets target tokens query covariate patch embeddings as keys and values through a gated residual branch. \emph{Global modulation} flattens the ordered covariate patch embeddings, maps them to one sequence-level condition, and combines it with diffusion time to generate AdaLN parameters shared across target positions. \emph{Joint attention} concatenates covariate and target tokens so they interact in one attention sequence. \emph{Variate attention} maintains separate target and covariate streams: each first applies attention along time, then joint attention across the target and covariate variates at each temporal position. All four variants inject diffusion time through AdaLN; the diagrams omit this common pathway for clarity.

\begin{table}[H]
\centering
\caption{Covariate-conditioning mechanisms on five EPF datasets with known-future covariates and $(W,H)=(168,24)$. DiTS uses Time-aligned Modulation; the alternatives use cross-attention, global modulation, joint attention, or variate attention. Entries are MSE and MAE (lower is better). Red boldface marks the best unrounded result for each dataset and metric.}
\label{tab:ablation_cond_appendix}
\small
\setlength{\tabcolsep}{3pt}
\resizebox{\textwidth}{!}{\begin{tabular}{l|cc|cc|cc|cc|cc|cc}
\toprule
Method & \multicolumn{2}{c|}{NP} & \multicolumn{2}{c|}{PJM} & \multicolumn{2}{c|}{BE} & \multicolumn{2}{c|}{FR} & \multicolumn{2}{c|}{DE} & \multicolumn{2}{c}{AVG} \\
\cmidrule(lr){2-3}\cmidrule(lr){4-5}\cmidrule(lr){6-7}\cmidrule(lr){8-9}\cmidrule(lr){10-11}\cmidrule(lr){12-13}
 & MSE & MAE & MSE & MAE & MSE & MAE & MSE & MAE & MSE & MAE & MSE & MAE \\
\midrule
DiTS & 0.160 & \textcolor{red}{\textbf{0.205}} & \textcolor{red}{\textbf{0.056}} & \textcolor{red}{\textbf{0.141}} & \textcolor{red}{\textbf{0.325}} & 0.232 & \textcolor{red}{\textbf{0.269}} & \textcolor{red}{\textbf{0.171}} & \textcolor{red}{\textbf{0.208}} & \textcolor{red}{\textbf{0.286}} & \textcolor{red}{\textbf{0.204}} & \textcolor{red}{\textbf{0.207}} \\
Cross attn. & 0.230 & 0.260 & 0.072 & 0.168 & 0.394 & 0.261 & 0.352 & 0.197 & 0.397 & 0.399 & 0.289 & 0.257 \\
Global mod. & 0.176 & 0.221 & 0.058 & 0.146 & 0.327 & 0.240 & 0.276 & 0.176 & 0.262 & 0.320 & 0.220 & 0.221 \\
Joint attn. & 0.164 & 0.210 & 0.057 & 0.141 & 0.334 & \textcolor{red}{\textbf{0.231}} & 0.291 & 0.172 & 0.230 & 0.304 & 0.215 & 0.212 \\
Variate attn. & \textcolor{red}{\textbf{0.158}} & 0.206 & 0.057 & 0.143 & 0.335 & 0.238 & 0.273 & 0.175 & 0.213 & 0.288 & 0.207 & 0.210 \\
\bottomrule
\end{tabular}}
\end{table}

\section{Showcases}
\begin{figure}[h]
    \centering
    \includegraphics[width=1\textwidth]{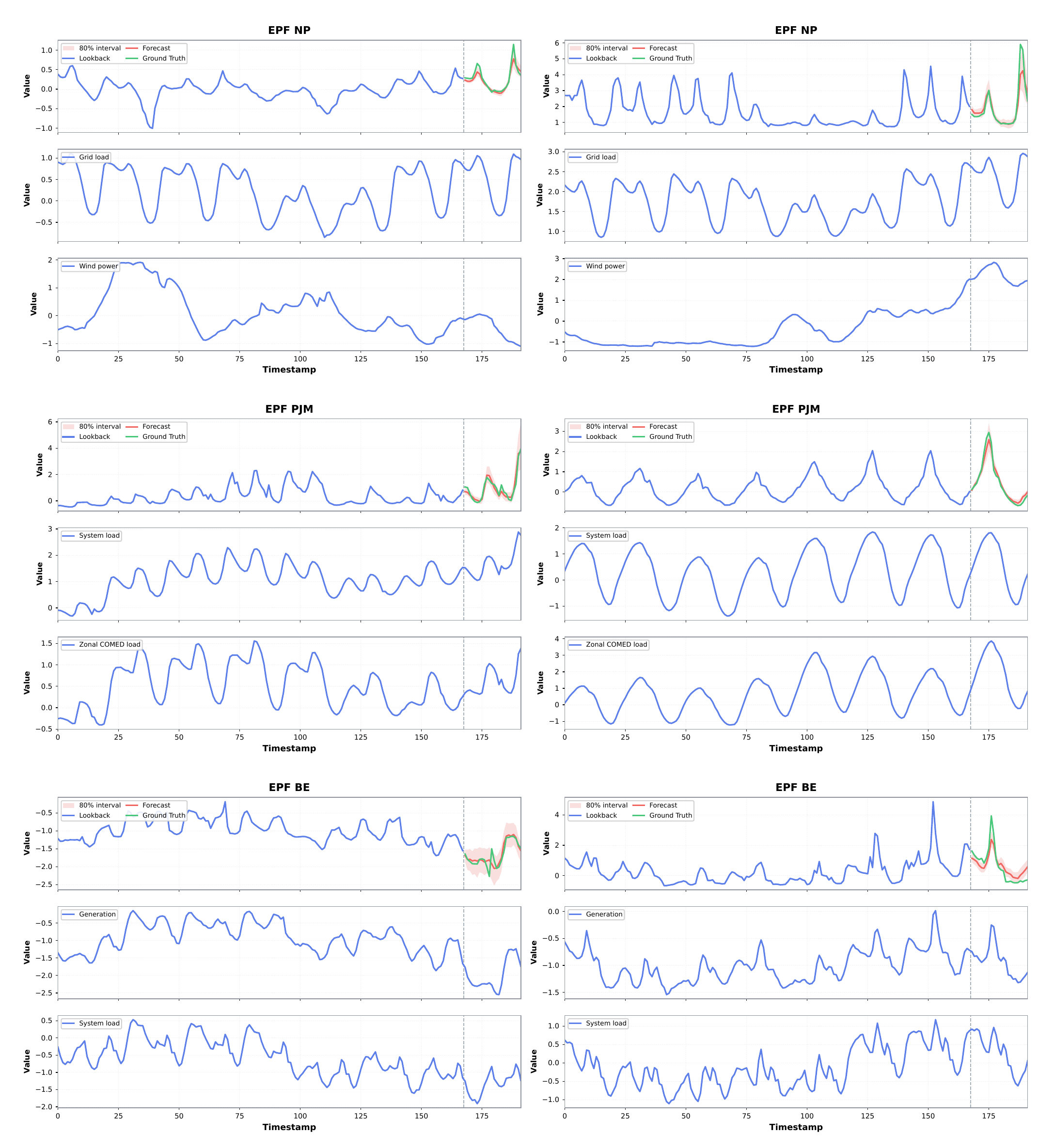}
    \caption{DiTS forecasts on EPF NP, PJM, and BE, with two examples from each dataset. In each example, the top panel shows the target and lower panels show available covariates. The vertical dashed line separates the observed history from the prediction horizon. Blue is the target history, red the mean of 10 sampled forecasts, green the ground truth, and the shaded band the 80\% prediction interval. Known-future covariates continue through the prediction horizon.}
    \label{fig:epf_showcase}
\end{figure}

\begin{figure}[h]
    \centering
    \includegraphics[width=1\textwidth]{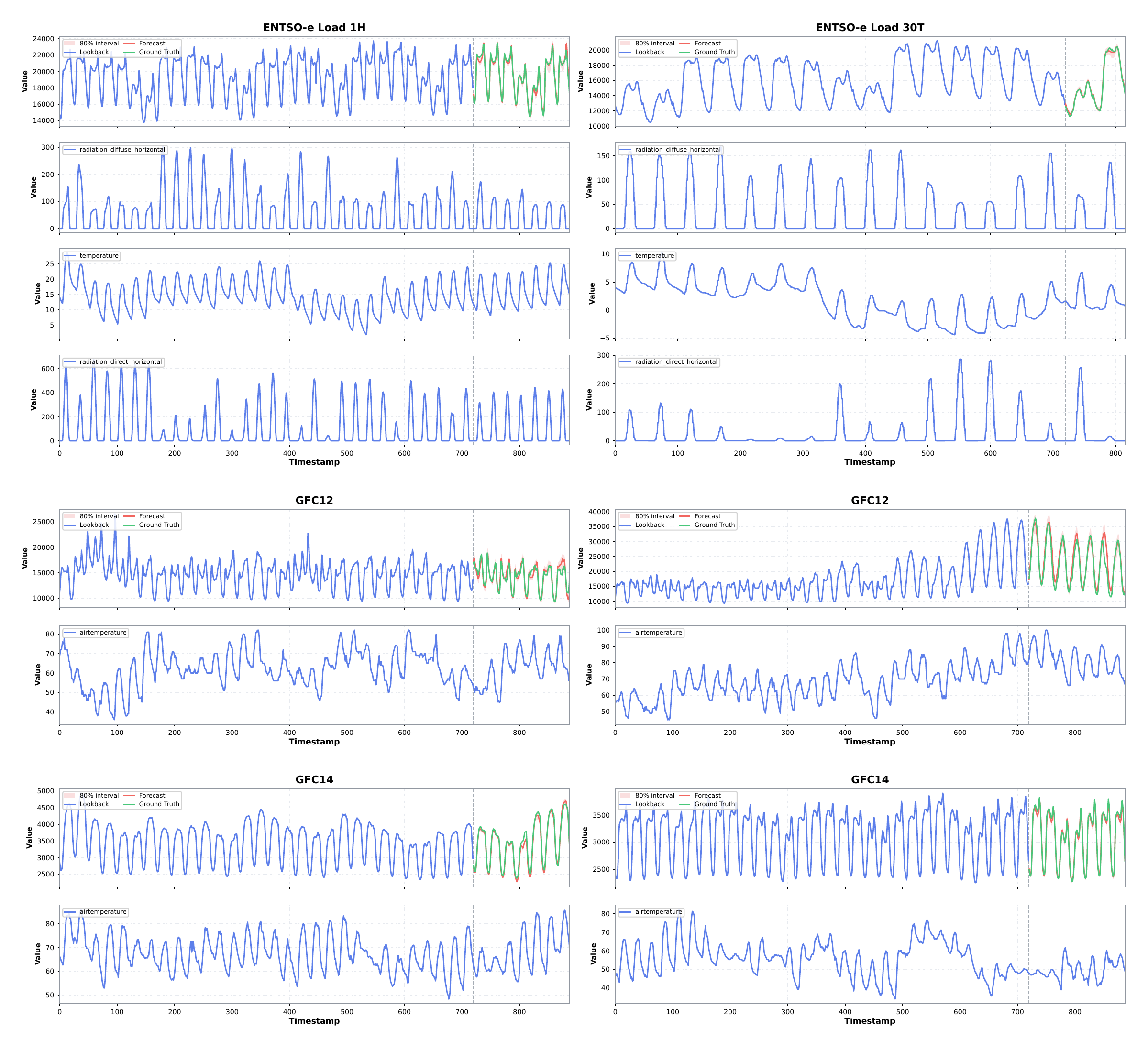}
    \caption{DiTS forecasts on selected electricity-load and GFC tasks: ENTSO-e Load (1H and 30T), GFC12, and GFC14. Each example displays the target above its covariates; the vertical dashed line marks the history--forecast boundary. Blue is the observed target, red the mean of 50 sampled forecasts, green the ground truth, and the shaded band the 80\% prediction interval. Covariates extend into the horizon only when known in advance.}
    \label{fig:fev_showcase}
\end{figure}

\begin{figure}[h]
    \centering
    \includegraphics[width=1\textwidth]{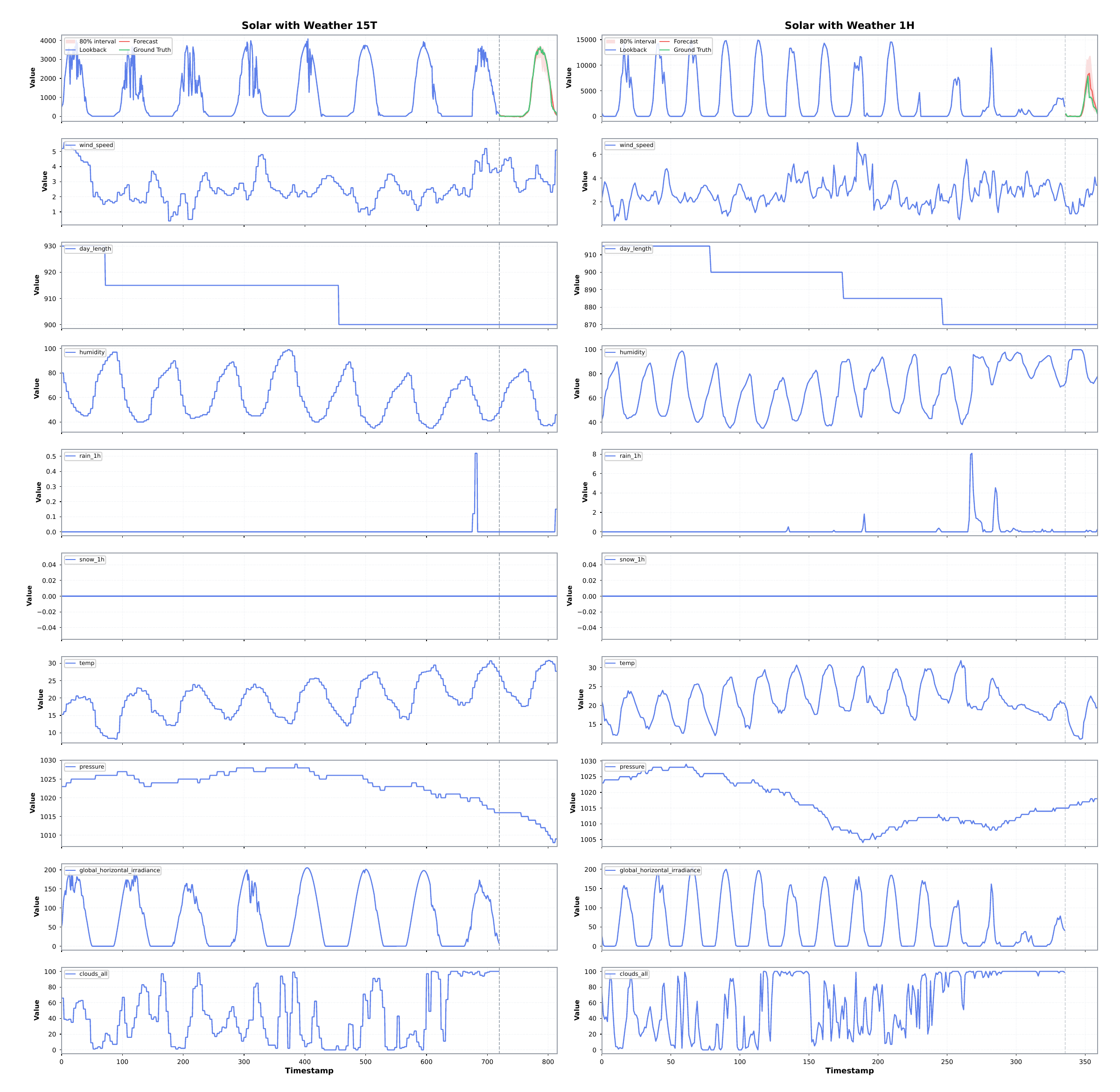}
    \caption{DiTS forecasts on the selected Solar with Weather tasks at 15-minute and hourly resolution. The target appears above the weather covariates, and the vertical dashed line separates history from the forecast horizon. Blue is the observed target, red the mean of 50 sampled forecasts, green the ground truth, and the shaded band the 80\% prediction interval. Weather covariates are shown in the horizon when available in advance; past-only covariates stop at the boundary.}
    \label{fig:solar_showcase}
\end{figure}

Figures~\ref{fig:epf_showcase}--\ref{fig:solar_showcase} show representative forecasts on EPF~\citep{wang2024timexer} and the selected probabilistic tasks. The forecast mean and 80\% prediction interval are estimated from 10 independently sampled trajectories for EPF and 50 for the selected tasks. Known-future covariates are shown through the prediction horizon, while covariates available only in the past end at the history--forecast boundary. 

\end{document}